\documentclass[11pt,a4paper]{article}
\usepackage[hyperref]{acl2020}
\usepackage{times}
\usepackage{latexsym,soul}

\usepackage{microtype}
\usepackage{caption}
\usepackage{subfigure}
\usepackage{graphicx}
\usepackage{makecell}

\aclfinalcopy % Uncomment this line for the final submission
\newcommand{\drc}[1]{\textrm{\color{purple} #1}}

\newcommand\name{\textsc{QuASE}}
\newcommand\quase[1]{\name{}$_{\small #1}$}
\newcommand\bert[1]{BERT$_{\small #1}$}
\newcommand\ignore[1]{}

\title{
\vspace*{-0.5in}
{{\small \hfill ACL'20}\\
\vspace*{.25in}} 
\name: Question-Answer Driven Sentence Encoding 
}
\author{Hangfeng He$^\dag$\qquad Qiang Ning$^\ddagger$\thanks{\, Part of this work was done while the author was at the University of Illinois at Urbana-Champaign.} \qquad Dan Roth$^\dag$ \\
  $^\dag$ University of Pennsylvania\qquad
  $^\ddagger$ Allen Institute for AI\\
  \texttt{hangfeng,danroth@seas.upenn.edu}\\\texttt{qiangn@allenai.org}}

\date{}

\begin{document}
\maketitle
\begin{abstract}
% Human annotations, especially those from experts, are costly for many natural language processing (NLP) tasks. One promising solution is to use natural language to annotate natural language. However, it remains an open problem how to get supervision signals or learn representations from natural language annotations. This paper studies the case where the annotations are in the format of question-answering (QA) and proposes {\em an effective way to learn useful representations} for other tasks. Specifically, we propose {\em QA-driven sentence encoder (\name) with two variants} for single-sentence tasks and sentence-pair tasks separately. We also find that the representation {\em retrieved from question-answer meaning representation (QAMR) dataset} can almost universally improve on a wide range of tasks in the low-resource setting, suggesting that such kind of natural language annotations indeed provide unique information on top of modern language representation models. This work may point out an alternative way to supervise NLP tasks.

Question-answering (QA) data often encodes essential information in many facets. This paper studies a natural question: Can we get supervision from QA data for other tasks (typically, non-QA ones)? For example, {\em can we use QAMR \cite{MSHDZ17} to improve named entity recognition?} 
We suggest 
%This paper suggests 
that simply further pre-training BERT is often not the best option, and propose the {\em question-answer driven sentence encoding (\name)} framework. \name{} learns representations from QA data, using 
%with the help of 
BERT or other state-of-the-art contextual language models. In particular, we observe the need to distinguish between two types of sentence encodings, depending on whether the target task is a single- or multi-sentence input; in both cases, the resulting encoding is shown to be an easy-to-use plugin for many downstream tasks. This work may point out 
an alternative way to supervise NLP tasks.\footnote{Our code and online demo are publicly available at \url{https://github.com/CogComp/QuASE}.}

%\hangfeng{A reviewer mention that QAMR is not illustrated before usage. I am also concerned about that. So I add a citation in the original version.}
%\dr{What if we add a reference?}
%\dr{Alternatively, we could get rid of this sentence, and add, "such as NER" in the previous sentence. Or, just leave it}

%\qn{my experience with editorial services is that the footnote should be after the punctuation}
% \hangfeng{the full description of QAMR first, or people don't know what's QAMR. Actually, I think a better example is to use QA-SRL to help SRL as in previous version.}
% \qn{I don't think we have space to explain QAMR, although I feel it's not even necessary, is it? People would infer, if they don't know already, that QAMR is a QA dataset.}
% \qn{I didn't want to leave an impression that we only learn from QA data that is similar to the target task, like QA-SRL and SRL.}
\end{abstract}

%!TEX root = root.tex
\section{Introduction}
\label{sec:intro}
% \qn{motivation, proposal, details of the proposal, results, importance of the work, relation to stilts/others, rest of the paper}

It is labor-intensive to acquire human annotations for NLP tasks which require research expertise. For instance, one needs to know thousands of semantic frames in order to provide semantic role labelings (SRL) \cite{PalmerGiXu10}. 
It is thus an important research direction to investigate how to get supervision signals from {\em indirect} data 
and
%as well as 
improve one's target task. 
% As these indirect data either have been collected for other research projects, or can be collected in a more efficient way, supervision signals from indirect data are often deemed cheap.
This paper studies the case of learning from question-answering (QA) data for other tasks (typically not QA). We choose QA because (1) a growing interest of QA has led to many large-scale QA datasets available to the community; (2) a QA task often requires comprehensive understanding of language and may encode rich information that is useful for other tasks; (3) it is much easier to answer questions relative to a sentence than to annotate linguistics phenomena in it, making this a plausible supervision signal~\cite{Roth17}.

% A promising approach to address this issue is to use natural language (NL) to annotate NL (i.e., {\em NL annotations}).
% Because annotators do not need to learn convoluted NLP concepts, we can often get quality NL annotations in large scale. Similar works along this line include natural logic for textual entailment (TE) \cite{maccartney2007natural}, QA-SRL \cite{HeLeZe15}, QAMR \cite{MSHDZ17}, and zero-shot relation extraction (RE) via reading comprehension \cite{levy-etal-2017-zero}. 

% A critical question is how to use NL annotations effectively.
% For instance, QA-SRL is NL annotations for SRL, but QA-SRL data are, in their surface form, very different from SRL; as Sec.~\ref{subsec:symbol-QA} shows, it is hard to learn a reasonably good SRL parser from QA-SRL data. We think that not only for SRL and QA-SRL, but it is generally hard to obtain reliable symbolic representations from NL annotations because of the high flexibility of NL.

There has been work showing that QA data for task $\mathcal{A}$ can help another QA task $\mathcal{T}$, conceptually by further pre-training the {\em same} model on $\mathcal{A}$ (an often larger) before training on $\mathcal{T}$ (a smaller) \cite{talmor-berant-2019-multiqa, sun-etal-2019-improving}.
However, it remains unclear how to use these QA data when the target task does not share the same model as the QA task, which is often the case when the target task is {\em not} QA.
For instance, QA-SRL \cite{HeLeZe15}, which uses QA pairs to represent those predicate-argument structures in SRL, should be intuitively helpful for SRL parsing, but the significant difference in their surface forms prevents us from using the {\em same} model in both tasks. 
% One may also suggest that we design rules to convert those QA pairs in QA-SRL to be the same format of SRL, but on one hand, the conversion may be too noisy to be helpful; on the other hand, this strategy fails when there is no obvious clue for the conversion (e.g., imagine converting QA-SRL data to named entity annotations).
% \hangfeng{I don't know why we need to convert QA-SRL data to NER in the situation where we want to use QA-SRL to help SRL.}

The success of modern language modeling techniques, e.g., ELMo \cite{peters-etal-2018-deep}, BERT \cite{devlin-etal-2019-bert}, and many others, has pointed out an alternative solution to this problem. That is, to further pre-train\footnote{We clarify three types of training: pre-training, further pre-training, and fine-tuning. Pre-training refers to the training of sentence encoders on unlabeled text; further pre-training refers to continuing training the sentence encoders on an intermediate, non target-task-specific labeled data (e.g. QA data); fine-tuning refers to training on the target task in the fine-tuning approach.} a neural language model (LM) on these QA data {\em in certain ways}, obtain a sentence encoder, and use the sentence encoder for the target task, either by fine-tuning or as additional feature vectors. We call this  general framework %\qn{change general to meta?} \dr{I think that ``general" is fine.}
{\em question-answer driven sentence encoding (\name)}. A straightforward implementation of \name{} is to first further pre-train BERT (or other LMs) on the QA data in the standard way, as if this QA task is the target, and then fine-tune it on the real target task. This implementation is technically similar to STILTS \cite{phang2018sentence}, except that STILTS is mainly further pre-trained on textual entailment (TE) data. 
% As STILTS only reported using textual entailment (TE) data to improve tasks in GLUE \qn{refs}, one contribution of this paper is more experimental results showing the benefit of using {\em QA} data on some tasks.

However, similar to the observations made in STILTS and their follow-up works \cite{wang2019can}, we find that additional QA data does not necessarily help the target task using the implementation above.
While it is unclear how to predict this behaviour
%when this happens in theory
, we do find that this happens a lot for tasks whose input is a single sentence, e.g., SRL and named entity recognition (NER), instead of a sentence pair, e.g., TE. This might be because QA is itself a paired-sentence task, and the implementation above (i.e., to further pre-train BERT on QA data) may learn certain attention patterns that can transfer to another paired-sentence task more easily than to a single-sentence task.
Therefore, we argue that, for single-sentence target tasks, \name{} should restrict the interaction between the two sentence inputs when it further pre-trains on QA data. We propose a new neural structure for this and name the resulting implementation s-\name{}, where ``s'' stands for ``single;'' in contrast, we name the straightforward implementation mentioned above p-\name{} for ``paired.'' Results show that s-\name{} outperforms p-\name{} significantly on $3$ single-sentence tasks---SRL, NER, and semantic dependency parsing (SDP)---indicating the importance of this distinction.

Let \quase{\mathcal{A}} be the \name{} further pre-trained on QA data $\mathcal{A}$.
We extensively compare 6 different choices of $\mathcal{A}$: TriviaQA \cite{joshi-etal-2017-triviaqa}, NewsQA \cite{trischler-etal-2017-newsqa}, SQuAD \cite{rajpurkar-etal-2016-squad}, relation extraction (RE) dataset in QA format (QA-RE for short) \cite{levy-etal-2017-zero}, Large QA-SRL \cite{fitzgerald-etal-2018-large}, and QAMR \cite{MSHDZ17}. Interestingly, we find that if we use s-\name{} for single-sentence tasks and p-\name{} for paired-sentence tasks, then \name{}${_\textrm{\small QAMR}}$ improves all 7 tasks\footnote{SRL, SDP, NER, RE, co-reference resolution (Coref), TE and machine reading comprehension (MRC).} in low resource settings, with an average error reduction rate of 7.1\% compared to BERT.\footnote{BERT is close to the state-of-the-art in all these tasks.}
While the set of tasks we experimented with here is non-exhaustive, we think that \name{}${_\textrm{\small QAMR}}$ has the potential of improving on a wide range of tasks.

This work has three important implications.
First, it provides supporting evidence to an important alternative to supervising NLP tasks: using QA to annotate language, which has been discussed in works such as QA-SRL, QAMR, and QA-RE. 
If it is difficult to teach annotators the formalism of a certain task, perhaps we can instead collect QA data that query the target phenomena and thus get supervision from QA for the original task (and possibly more).
% Because annotators do not need to learn convoluted NLP concepts, we can often get quality NL annotations in large scale.
Second, the distinction between s-\name{} and p-\name{} suggests that sentence encoders should consider some properties of the target task (e.g., this work distinguishes between single- and multi-sentence tasks). 
% \dr{I don't understand the role of the following sentence here; can we get rid of it?}
% For instance, this work studies the impact of the number of input sentences, but it poses an important research question about whether and when supervision signals can be transferred between tasks.
Third, the good performance of \quase{QAMR} suggests that predicate-argument identification is an important capability that many tasks rely on; in contrast, many prior works observed that only language modeling would improve target tasks generally.

\section{QA Driven Sentence Encoding}
% \qn{QA-driven or QA Driven?} \dr{looks good to me} \hangfeng{I think QA Driven is better.}
\label{sec:model}

\begin{figure*}
		\centering
		\subfigure[s-\name{} for single-sentence tasks]{
			\centering
			\includegraphics[scale=0.26]{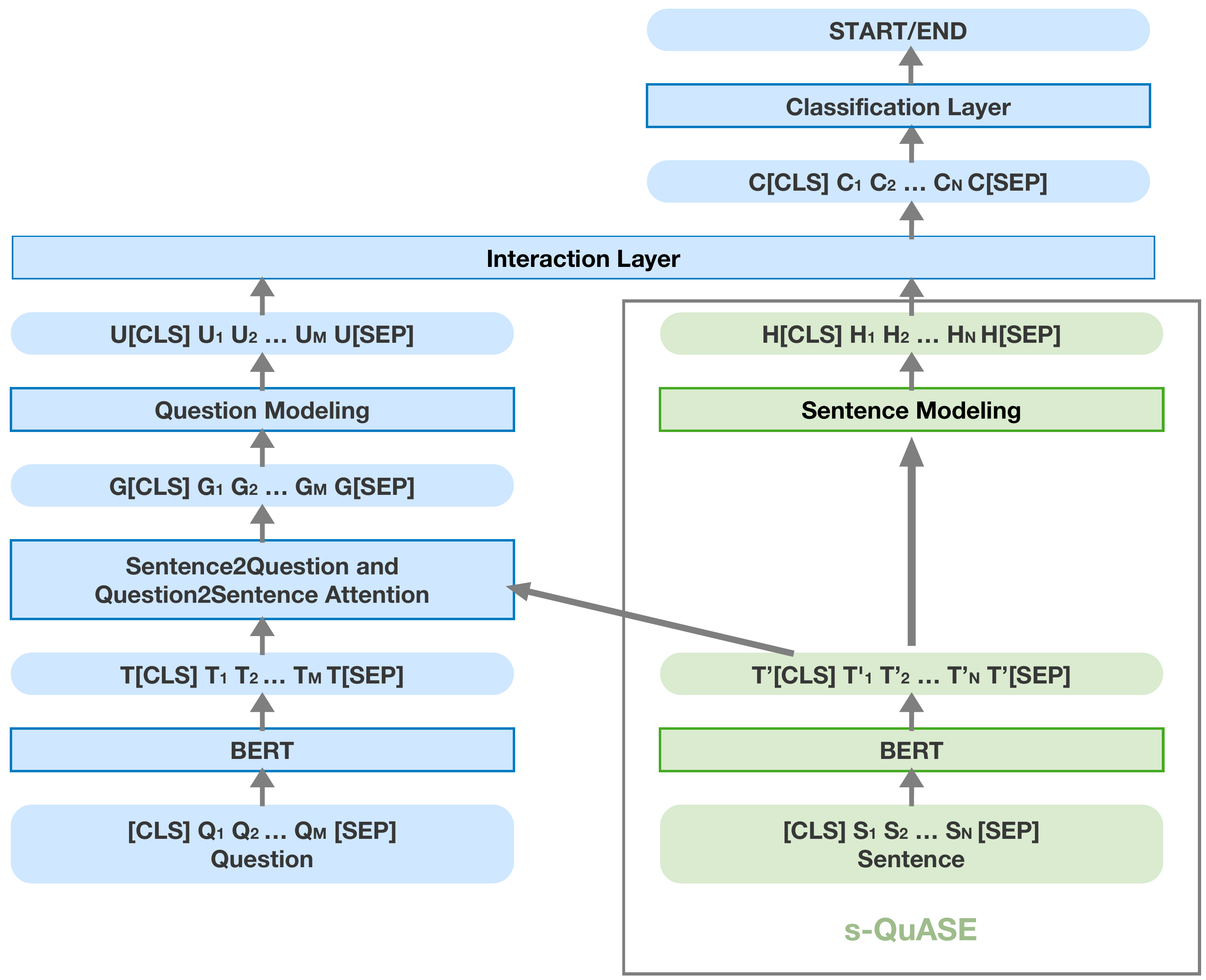}
			\label{fig:semanticbert-arch}}
        \hspace{0.2in}
        \subfigure[p-\name{} for paired-sentence tasks]{
		\centering
		\includegraphics[scale=0.23]{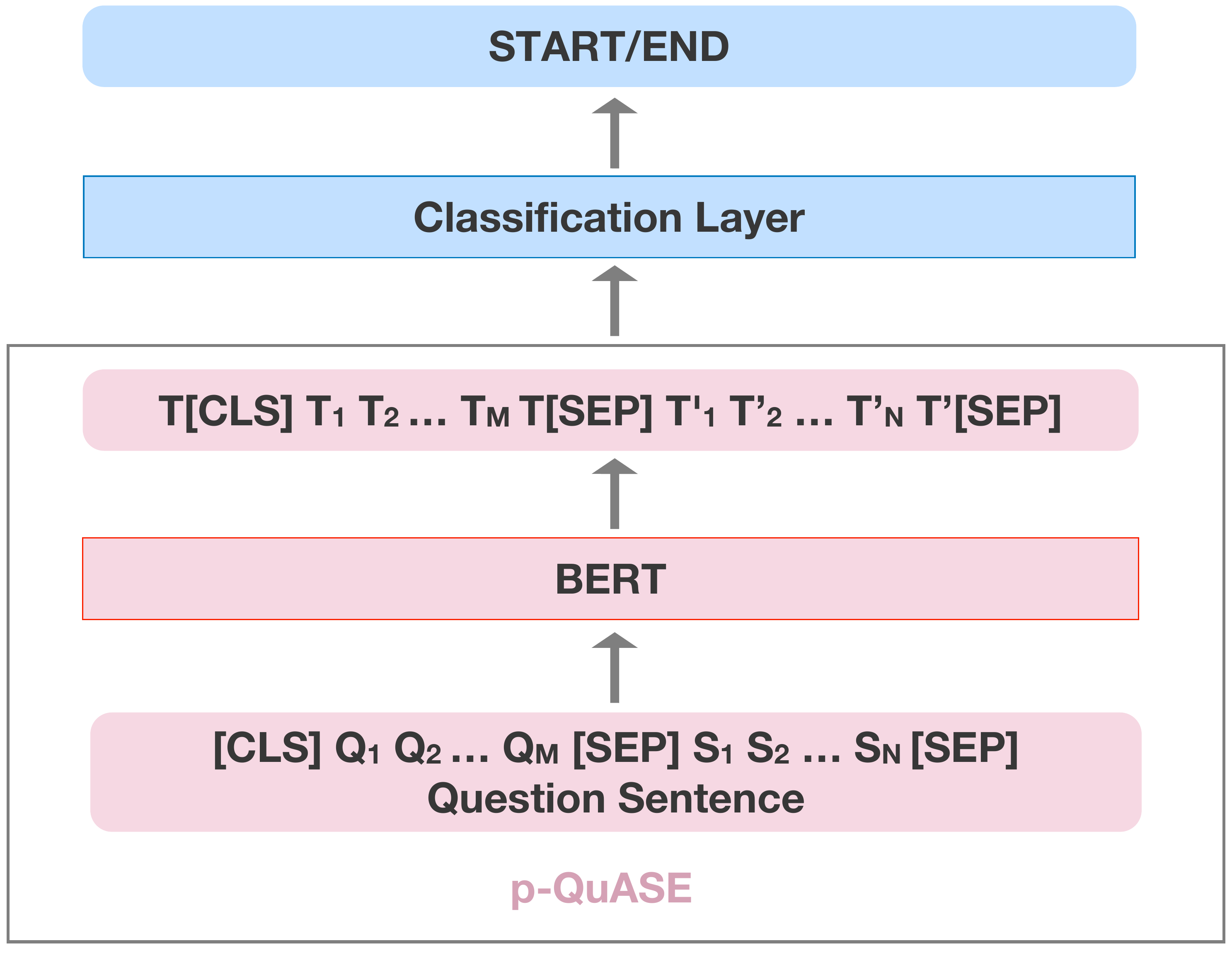}
			\label{fig:conditional-semanticbert-arch}}
        \hspace{0.01in}
        \caption{ 
        Two implementations of \name{}: s-\name{} for single-sentence tasks, and p-\name{} for paired-sentence tasks. Both structures are further pre-trained on QA data, and the parts in the black boxes are used by target tasks. While p-\name{} is the standard way of fine-tuning BERT
        %(by concatenating the sentence and question using special tokens)
        , s-\name{} restricts the interaction between the sentence and the question. Specifically, the sentence encodings in s-\name{} does not depend on the existence of the question. More details are given in Sec.~\ref{subsec:two-quase} and Appendix \ref{sec:quase-details} (including experimental settings in \ref{subsec:quase-experiments}, error analysis in \ref{subsec:erro-analysis} and ablation analysis in \ref{subsec:quase-ablation}). 
        % The architectures of s-\name{} and p-\name{} for \qnch{further} pre-training on QA data. 
        % Note that p-\name{} has the same architecture as BERT. 
        % As for s-\name{}, we have a sentence modeling layer on the top of BERT component for the sentence representation. Similarly, we add a Sentence2Question and Question2Sentence Attention layer before question modeling layer on the top of BERT component for the question representation. After that, the interaction layer is used to learn the interaction between the question and the sentence, and its output is sent to the classification layer for MRC pre-training.  Those parts in black boxes are the components for applications. More details can be found in Appendix \ref{sec:quase-details}.
        %\qn{if p-\name{} is the same with finetuning BERT, maybe mention it explicitly?}
        % \qn{replace jpg to be pdf}
        % \qn{I remember one reviewer complained that this figure isn't very clear? At least to me, it's a bit too small; and since it's only on the second page, people won't understand the figure at all before reading the text. You can either put the figure later, or give a short description in the caption here.}
        }
		\label{fig:architectures}
\end{figure*}

% \qn{straightforward method: convert from QA to the target task; finetune and continue finetune when the target task is QA;  finetune and continue finetune when the target task is not QA; intuition why that won't always work, especially for single-sentence tasks; our proposal.}

This work aims to find an effective way to use readily available QA data to improve a target task that is typically not QA.
A natural choice nowadays---given the success of language models---is to further pre-train sentence encoders, e.g. BERT, on QA data {\em in certain ways}, and then use the new encoder in a target task. This general framework is called \name{} in this work, and the assumption is that the sentence encoders learned from QA data have useful information for the target task.

A straightforward implementation of \name{} is to further pre-train BERT on QA data {\em in the standard way}, i.e., fine-tune BERT as if this QA dataset is the target task, and then fine-tune BERT on the real target task.
% Similar ideas have been studied in STILTS \cite{phang2018sentence} and its follow-ups \qn{refs} (which we will discuss in Sec.~\ref{}).
However, we find that this straightforward implementation is less effective or even negatively impacts target tasks with single-sentence input; similar observations were also made in STILTS \cite{phang2018sentence} and its follow-ups \cite{wang2019can}: They further pre-train sentence encoders, e.g., ELMo, BERT, and GPT \cite{radford2018improving}, on TE data and find that it is not effective for the syntax-oriented CoLA task and the SST sentiment task in GLUE, which are both single-sentence tasks \cite{wang-etal-2018-glue}.

One plausible reason is that the step of further pre-training on QA data does not take into account some properties of the target task, for instance, the number of input sentences.
QA is inherently a paired-sentence task; a typical setup is, given a context sentence and a question sentence, predict the answer span. Further pre-training BERT on QA data will inevitably learn how to attend to the context given the question. This is preferable when the target task is also taking a pair of sentences as input, while it may be irrelevant or harmful for single-sentence tasks.
It points out that we may need two types of sentence encodings when further pre-training BERT on QA data, depending on the type of the target task.
The following subsection discusses this issue in detail.

% A straightforward method is to further pre-train sentence encoders (e.g. BERT) on QA data to get a better sentence encoder. We first discuss the issues of this straightforward method in Section \ref{subsec:continuing-training-SE}. Based on those issues, we propose to rethink the sentence representations and distinguish two types sentence representations in Section \ref{subsec:two-representations}. Finally, in Section \ref{subsec:two-quase}, \name{} with two variants are proposed for sentence tasks and sentence-pairs separately. 

% To take advantage of QA pairs for downstream tasks, a straightforward method is to continue training sentence encoders (e.g. BERT) on QA pairs. 
% As we show later in Sec. \ref{subsec:necessarity-two-\name{}}, we find that BERT continuously trained with QAMR seems improve less or even  doesn't improve on single-sentence tasks compared to sentence-pair tasks. This is consistent with the finding in \cite{phang2018sentence}. They find that compared to single-sentence tasks, sentence-pair tasks seem to benefit more from sentence encoders (i.e. BERT, GPT \cite{radford2018improving} and ELMo) that has been additionally trained on TE data. It points out that we may need to distinguish the sentence representations for single-sentence pairs and sentence-pair tasks.

\subsection{Two Types of Sentence Encodings}
\label{subsec:two-representations}

Standard sentence encoding is the problem of converting a sentence  $S$=[$w_1$, $w_2$, $\cdots$, $w_n$] to a sequence of vectors $h(S)$=[$h_1$, $h_2$, $\cdots$, $h_n$] (e.g., skip-thoughts \cite{kiros2015skip}). 
Ideally, $h(S)$ should encode {\em all} the information in $S$, so that it is task-agnostic: given a target task, one can simply probe $h(S)$ and retrieve relevant information.
In practice, however, only the information relevant to the training task of $h(S)$ is kept. 
For instance, when we have a task with multi-sentence input (e.g., QA and TE), the attention pattern $A$ among these sentences will affect the final sentence encoding, which we call $h_A(S)$; in comparison, we denote the sentence encoding learned from single-sentence tasks by $h(S)$, since there is no cross-sentence attention $A$. In a perfect world,  the standard sentence encoding $h(S)$ expresses also the
%includes 
conditional sentence encoding $h_A(S)$.  However, we believe that there is a trade-off between the quality and the quantity of semantic information 
%that 
a model can encode.
%in practice. 
Our empirical results
%later 
corroborate this conclusion and more details can be found in Appendix \ref{subsec:erro-analysis}.

The distinction between the
%of two types of 
sentence encodings types may explain the negative impact of using QA data for some single-sentence tasks: Further pre-training BERT on QA data essentially produces a sentence encoding with cross-sentence attentions $h_A(S)$, while the single-sentence tasks expect $h(S)$.
These two sentence encodings may be very different: One view is from the theory of information bottleneck \cite{tishby2000information,tishby2015deep}, which argues that training a neural network on a certain task is extracting an approximate {\em minimal} sufficient statistic of the input sentences with regard to the target task; information irrelevant to the target task is {\em maximally} compressed. In our case, this corresponds to the process where the conditional sentence encoding compresses the information irrelevant to the relation, which will enhance the quality but reduce the quantity of the sentence information.

\subsection{Two Implementations of \name{}}
\label{subsec:two-quase}

In order to fix this issue, we need to know how to learn $h(S)$ from QA data. However, since QA is a paired-sentence task, the attention pattern between the context sentence and the question sentence is important for successful further pre-training on QA. Therefore, we propose that if the target task is single-sentence input, then further pre-training on QA data should also focus on single-sentence encodings in the initial layers; the context sentence should not interact with the question sentence until the very last few layers.
This change is expected to hurt the capability to solve the auxiliary QA task, but it is later proved to transfer better to the target task.
This new treatment is called s-\name{} with ``s'' representing ``single-sentence,'' while the straightforward implementation mentioned above is called p-\name{} where ``p'' means ``paired-sentence.''
The specific structures are shown in Fig.~\ref{fig:architectures}.

\subsubsection{s-\name{}}

The architecture of s-\name{} is shown in Fig.~\ref{fig:semanticbert-arch}. 
When further pre-training it on QA data, the context sentence and the question sentence are fed into two pipelines.
We use the same Sentence2Question and Question2Sentence attention as used in BiDAF \cite{seo2016bidirectional}.
Above that, ``Sentence Modeling,'' ``Question Modeling,'' and ``Interaction Layer'' are all bidirectional transformers \cite{vaswani2017attention} with 2 layers, 2 layers, and 1 layer, respectively.
% the ``Sentence Modeling'' layer and the ``Question Modeling'' layer are both two-layer bidirectional transformers, while the ``Interaction Layer'' is a single-layer bidirectional transformer. 
Finally, we use the same classification layer as BERT, which is needed for training on QA data. 
Overall, this implementation restricts interactions between the paired-sentence input, especially from the question to the context, because when serving the target task, this attention will not be available.
%More details of \name{} can be found in Appendix \ref{subsec:\name{}-details}.
%\qn{I don't know what is going to be in that appendix}

% Tenney et al.~\shortcite{tenney2019bert} point out that lower layers of BERT encode more local syntax, while higher layers capture more complex semantics. This finding is consistent with our intuition, and we add a sentence modeling layer on top of BERT to include more complex semantics.
% \qn{I don't think Tenney's work can speak for us. If you think it's critical, let me know.}
% \paragraph{\textbf{Using s-\name{} in target tasks.}}
\noindent\textbf{Using s-\name{} in target tasks.}
Given a sentence $S$, s-\name{} can provide a sequence of hidden vectors $h(S)$, i.e., the output of the ``Sentence Modeling'' layer in Fig.~\ref{fig:semanticbert-arch}.
Although $h(S)$ does not rely on the question sentence, $h(S)$ is optimized so that upper layers can use it to handle those questions in the QA training data, so $h(S)$ indeed captures information related to the phenomena queried by those QA pairs.
For single-sentence tasks, we use $h(S)$ from s-\name{} as additional features, and concatenate it to the word embeddings in the input layer of any specific neural model.\footnote{ We mainly use concatenation in both types of \name{}. However, we also use replacement in some experiments and we will note these cases later in this paper.}

\subsubsection{p-\name{}}

The architecture of p-\name{} is shown in Fig.~\ref{fig:conditional-semanticbert-arch}, which is the standard way of pre-training BERT.
That is, when further pre-training it on QA data, the context sentence and the question sentence form a single sequence (separated by special tokens) and are fed into BERT.

% \paragraph{\textbf{Using p-\name{} in target tasks.}}
\noindent\textbf{Using p-\name{} in target tasks.}
Given a sentence pair S (concatenated), p-\name{} produces $h_A(S)$, i.e., the output of the BERT module in Fig.~\ref{fig:conditional-semanticbert-arch}.
One can of course continue fine-tuning p-\name{} on the target task, but we find that adding p-\name{} to an existing model for the target task is empirically better (although not very significant); specifically, we try to add $h_A(S)$ to the final layer before the classification layer, and we also allow p-\name{} to be updated when training on the target task, although it is conceivable that other usages may lead to even stronger results.
For instance, when the target task is token classification, e.g., MRC, we can simply concatenate the vectors of $h_A(S)$ at each timestamp to any existing model; when the target task is sentence classification, e.g., TE, we apply max-pooling and average-pooling on $h_A(S)$, respectively, and concatenate the two resulting vectors to any existing model before the final classification layer.

% We use BERT to pre-train on QA dataset and get p-\name{} as shown in Fig.~\ref{fig:conditional-semanticbert-arch} for sentence-pair tasks. Given a sentence $S$ = [$w_1$, $w_2$, $\cdots$, $w_n$] based on an attention $A$ = [$q_1$, $q_2$, $\cdots$, $q_m$], our p-\name{} can provide a distributed representation $h(S|A)$ = [$h_1$, $h_2$, $\cdots$, $h_n$]. we add p-\name{} to the layer before the classification layer, and fine-tune it on downstream tasks.
% \qn{I'm confused. Isn't p-\name{} the same as further pre-training BERT on QA and then fine-tuning on the target task?}

% {\bf Token Classification Tasks.} In this case, our p-\name{} provides a conditional sentence representation of sentence $S$ with the attention $A$ as $h(S|A)$ = [$h_1, h_2, \cdots, h_n$]. For each token in the sentence, the hidden vector $h_t$ will be concatenated to the original BERT token embeddings before the classification layer. We can see that MRC relies on conditional token embeddings ($h(S|A)$) of the paragraph ($S$) given the question ($A$).

% {\bf Sentence Classification Tasks.} Similarly, in this case, our p-\name{} will provide a conditional sentence representation for sentence $S$ with the attention $A$ as $h(S|A)$ = [$h_1, h_2, \cdots, h_n$]. We first use the max pool and average pool to get related hidden vectors $h_{max}$ and $h_{avg}$, and then we concatenate these two vectors to the original BERT sentence embedding (the embedding of [CLS]) before the classification layer. TE relies on conditional sentence embedding of the hypothesis ($S$) given the premise ($A$). 

\subsection{Related Work on Sentence Encoding 
%\qn{make this a separate section?}
}
\label{subsec:other-SEs}
%To show whether \name{} can also provide extra features than other STOA embeddings\footnote{The reported STOA models for SRL and Coref is based on ELMo embeddings and the reported STOA model for NER is based on Flair embeddings.}, such as ELMo and Flair, we compare s-\name{} embeddings with ELMo embeddings on SRL and Coref, and compare s-\name{} embeddings with Flair on NER. The results are shown in Table. We find that \name{} has a better performance than ELMo and Flair, especially in the low-resource setting, which indicates that \name{} can provide extra features than ELMo and Flair.
Modern LMs are essentially sentence encoders pre-trained on unlabeled data and they outperform early sentence encoders such as skip-thoughts \cite{kiros2015skip}. While an LM like BERT can handle lexical and syntactic variations quite well, it still needs to learn from some annotations to acquire the ``definition'' of many tasks, especially those requiring complex semantics \cite{tenney2019you}.
Although we extensively use BERT here, we think that the specific choice of LM is orthogonal to our proposal of learning from QA data. Stronger LMs, e.g., RoBERTa \cite{liu2019roberta} or XLNet \cite{yang2019xlnet}, may only strengthen the proposal here. 
This is because a stronger LM represents \textit{unlabeled} data better, while the proposed work is about how to represent \textit{labeled} data better.

CoVe \cite{mccann2017learned} is another attempt to learn from indirect data, translation data specifically. However, it does not outperform ELMo or BERT in many NLP tasks  \cite{peters-etal-2018-deep} and probing analysis \cite{tenney2019you}.
In contrast, our \name{} will show stronger experimental results than BERT on multiple tasks.
In addition, we think QA data is generally cheaper to collect than translation data.

% Previous unsupervised LMs (ULMs), such as ELMo and BERT, do not perform well in tasks that require complex semantics \cite{tenney2019you}.
% Although CoVe \cite{mccann2017learned} is trained on translation signals, it is significantly worse than Elmo and BERT in helping NLP tasks \cite{peters-etal-2018-deep} and probing analysis \cite{tenney2019you}. 
% Our \name{} is able to provide extra information that BERT doesn't include on some tasks that requires complex semantics. \name{} precedes these ULMs by making use of low-cost signals, and it is orthogonal to other ULMs (e.g., we can have a similar \name{} for XLNet \cite{yang2019xlnet}). 

The proposed work is highly relevant to \citet{phang2018sentence} and their follow-up works \cite{wang2019can}, which use further pre-training on data-rich intermediate supervised tasks and aim to improve another target task.
The key differences are as follows:
%We mainly have the following differences.
First, we distinguish two types of sentence encodings, which provide explanation to their puzzle that sentence-pair tasks seem to benefit more from further pre-training than single-sentence tasks do. 
Second, they
%\citet{phang2018sentence} and their follow-up works \cite{wang2019can} 
only focus on fine-tuning based methods which cannot be easily plugged in many single-sentence tasks such as SRL and Coref, while we analyze both fine-tuning based and feature-based approaches. 
Third, they mainly use TE signals for further pre-training, and evaluate their models on GLUE \cite{wang-etal-2018-glue} which is a suite of tasks very similar to TE. Our work instead makes use of QA data to help tasks that are typically not QA.
Fourth, from their suite of further pre-training tasks, they observe that only further pre-training on language modeling tasks has the power to improve a target task in general, while we find that QAMR may also have this potential, indicating the universality of predicate-argument structures in NLP tasks. 
% Phang et al.~\shortcite{phang2018sentence} propose to use a second stage of pre-training with data-rich intermediate supervised tasks. There are several difference between ours and theirs. 
% First, we distinguish two types of sentence representations, which illustrates their puzzle that sentence-pair tasks seem to benefit more from BERT that has been additionally trained on TE data than single-sentence tasks. Second, they only focus on fine-tuning based methods which cannot be easily used in many single-sentence tasks such as SRL and Coref, but we analyze both fine-tuning based and feature-based methods. Another difference is that they mainly use TE signals for supplementary training, and evaluate their models on GLUE which are similar to TE. Our work instead makes use of QA signals to help seven quite different tasks.
%\qn{Should we include Sent BERT?} \dr{currently we have a bit of space, so a sentence on this would be good (there are two versions, from Iryna Gurevitch and AI2)}

Our work is also related to Sentence-BERT \cite{reimers2019sentence} in terms of providing a better sentence representation. However, their focus was deriving semantically meaningful sentence embeddings that can be compared using cosine-similarity, which reduces the computational cost of finding the most similar pairs. In contrast, \name{} provides a better sentence encoder in the same format as BERT (a sequence of word embeddings) to better support tasks that require complex semantics.
\section{Applications of \name{}
%\qn{applications of \name{}?}
}
\label{sec:experiments}
\begin{table}
\centering
\scalebox{0.86}{
\begin{tabular}{c||c|c||c|c}
\Xhline{2\arrayrulewidth}
&\multicolumn{2}{c||}{Single-sentence} & \multicolumn{2}{c}{Paired-sentence} \\ \hline
System & SRL & RE & TE & MRC\\ \hline
% Split & 10\% & 10\% & 10\% & 10\% \\ \hline
BERT & {\bf 34.17} & {\bf 62.99}& 78.29 & 79.90 \\ \hline
\bert{QAMR} & 32.92& 50.16& {\bf 78.73} & {\bf 82.96}\\ \Xhline{2\arrayrulewidth}
\end{tabular}}
\caption{
The naive way of training BERT on QAMR (\bert{QAMR}) negatively impacts single-sentence tasks.
We only use 10\% training data for simplicity. 
We use BERT/\bert{QAMR} to produce feature vectors for a BiLSTM model (SRL) and a CNN model (RE);
for TE and MRC, we fine-tune BERT/\bert{QAMR}.
% Comparison between BERT and BERT(QAMR) (additional training of BERT on QAMR) on two single-sentence tasks (SRL and RE) and two paired-sentence tasks (TE and MRC). For simplicity, we only use $10\%$ training examples for all four tasks \qn{remove the 10\% row in the table; also say in the caption sth like: these numbers may differ from other tables because of what}. We use BERT and BERT(QAMR) in a feature-based approach for single-sentence tasks and in a fine-tuning approach for paired-sentence tasks. Following standard practice, we use a simple BiLSTM model for SRL and a simple CNN model for RE.
%More experimental settings in Appendix \ref{sec:experiment-details}. 
% \qn{highlight what message you want to say here; e.g., BERT (QAMR) is under BERT for single-sentence tasks, but I don't see this verbiage in the current caption.}
}
\label{table:simple-additional-training}
\end{table} 
\begin{table*}
\centering
\scalebox{0.84}{
\begin{tabular}{c||c|c||c|c||c|c||c|c||c|c}
\Xhline{2\arrayrulewidth}
 & \multicolumn{6}{c||}{Single-Sentence Tasks} & \multicolumn{4}{c}{Paired-Sentence Tasks} \\ \hline
Tasks & \multicolumn{2}{c||}{SRL} & \multicolumn{2}{c||}{SDP} &  \multicolumn{2}{c||}{NER} &  \multicolumn{2}{c||}{TE} &  \multicolumn{2}{c}{MRC} \\ \hline
Split & 10\% & 100\% & 10\% & 100\% & 10\% & 100\% & 10\% & 30\% & 10\% & 100\% \\ \hline
s-\name{} & {\bf 46.42} & {\bf 70.13} & {\bf 76.08} & {\bf 87.29} & {\bf 70.69} & {\bf 87.10} & 52.25 &57.30 &44.67 & 67.09\\ \hline
p-\name{} & 32.92 & 66.40 & 70.92& 86.43 & 49.97 & 85.23 & {\bf 57.29}& {\bf 60.49} & {\bf 48.29} & {\bf 72.97}\\ \Xhline{2\arrayrulewidth}
\end{tabular}}
\caption{
Probing results of the sentence encoders from s-\name{} and p-\name{}.
In all tasks, we fix the model \name{} and use the sentence encodings as input feature vectors for the model of each task.
In order to keep the model structure as simple as possible, we use BiLSTM for SRL, NER, and TE, Biaffine for SDP, and BiDAF for MRC. %As for the split of training set, 
%In addition, 
We compare on 10\% and 100\% of the data in all tasks except TE, where we use 30\% to save run-time.
% We compare s-\name{}$_{QAMR}$ with p-\name{}$_{QAMR}$ on three single-sentence tasks (SRL, SDP and NER) and two paired-sentence tasks (TE and MRC). Note that we use s-\name{} and p-\name{} in the feature-based approach for all five tasks here, because s-\name{} can only be easily used in this approach. Following standard practice, we use a simple BiLSTM model for SRL, a simple Biaffine model for SDP, a simple BiLSTM model for NER, a simple BiLSTM model for TE, and BiDAF for MRC. 
% We only consider $30\%$ rather than $100\%$ of training examples for TE because the model is too time-consuming. 
%More details can be found in Appendix \ref{sec:experiment-details}. 
}
\label{table:s-QASE-vs-p-QASE}
\end{table*}
\begin{figure*}[t]
		\centering
		\subfigure[s-\name{}]{
			\centering
			\includegraphics[scale=0.30]{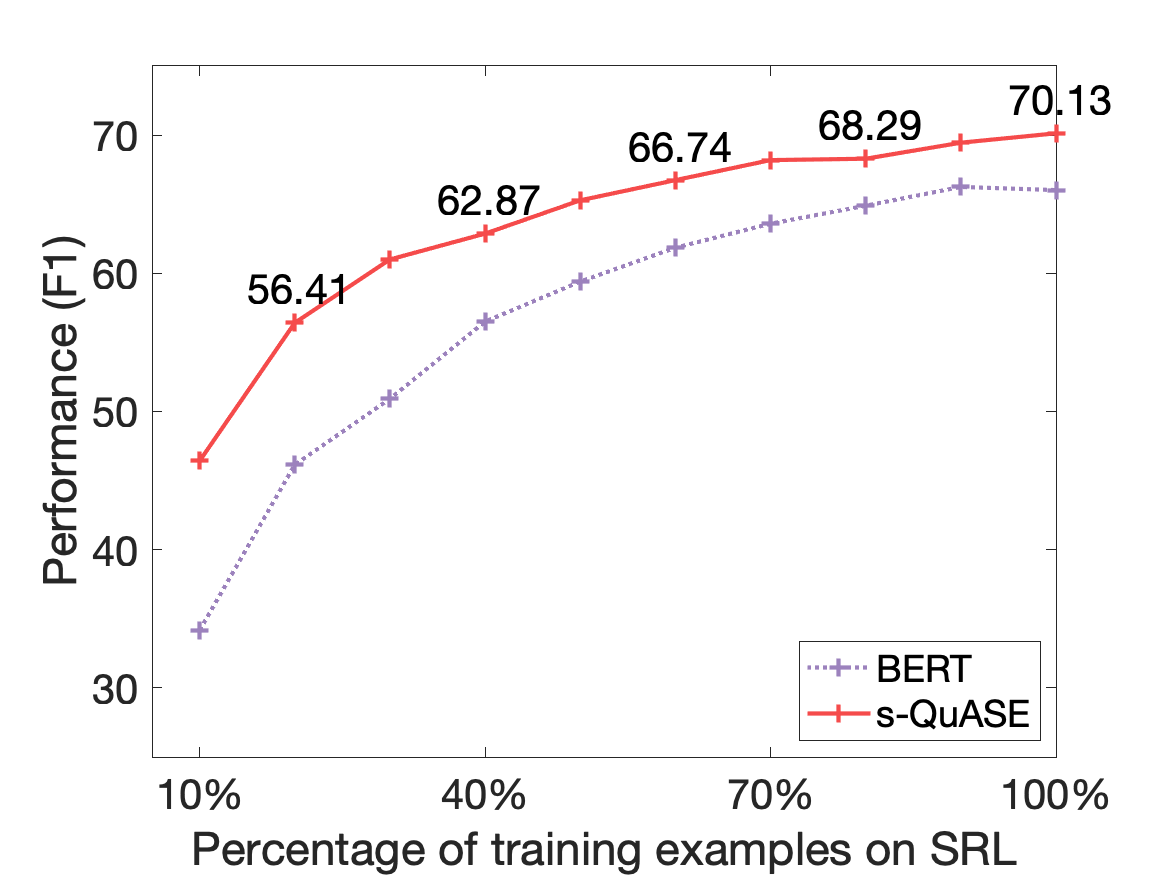}
			\label{fig:bert-qase-srl}}
     	\subfigure[p-\name{}]{
			\centering
			\includegraphics[scale=0.30]{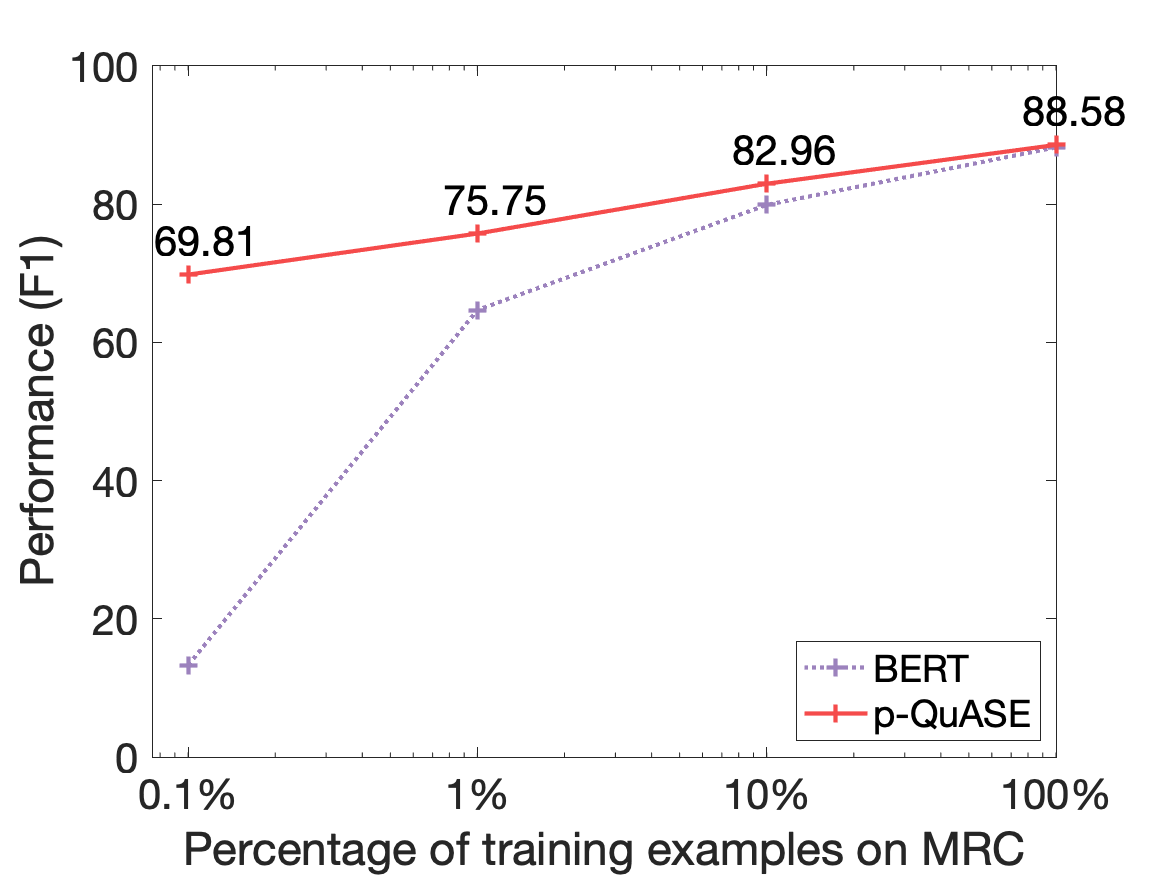}
			\label{fig:bert-qase-mrc}}
		\caption{ Sample complexity analysis of using BERT and \name{} on SRL and MRC. We find that much fewer training examples are needed with the help of \quase{QAMR}: with 50\% SRL training data, s-\name{} can achieve comparable performance as BERT trained on 100\%; with 0.1\% training data for MRC, p-\name{} can achieve a reasonably good performance of 69.81\%.
% 		Comparison between BERT and \name{}$_{QAMR}$ on SRL and MRC with different percentages of training data. For SRL, BERT and \name{} are used in the feature-based approach with a simple BiLSTM model. We use BERT and \name{} in the fine-tuning approach for MRC. 
		%More details can be found in Appendix \ref{sec:experiment-details}. 
% 		\qn{1. add several numbers (not necessarily all) to (a) as you did for (b). 2. make the fontsize of markers larger; they are still unnecessarily small. 3. I deleted model details to save space; we point to the appendix after all.}
}
\label{fig:sample-complexity}
\end{figure*}
\begin{table*}[t]
\centering
\scalebox{0.84}{
\begin{tabular}{c||c|c||c|c||c|c||c|c||c|c||c|c}
\Xhline{2\arrayrulewidth}
Models & \multicolumn{8}{c||}{s-\name{}} & \multicolumn{2}{c||}{p-\name{}} &
\multicolumn{2}{c}{}
\\ \hline 
 Tasks & \multicolumn{2}{c||}{SRL} & \multicolumn{2}{c||}{SDP} & \multicolumn{2}{c||}{NER} & \multicolumn{2}{c||}{RE} & \multicolumn{2}{c||}{TE} & \multicolumn{2}{c}{Avg} \\ \hline
Split & small & full & small & full & small & full & small & full & small & full & small & full\\ \hline \hline
BERT &34.17 & 66.02 & 75.49 & 90.13& 88.89& 91.38& 71.48& 86.33& 78.29& 84.09& 69.66& 83.59\\ %\hline
\name{}  & {\bf 50.16} & {\bf 72.59} & {\bf 78.30} & {\bf 90.78} & {\bf 90.64} & {\bf 92.16} & {\bf 77.14} & {\bf 86.80} & {\bf 78.94} & {\bf 84.97} & {\bf 75.04} & {\bf 85.46}\\ \hline \hline
TriviaQA &17.75 &39.69 &77.29 &90.43 &89.74 &91.70 &75.41 & {\bf 86.80}  & 78.50 & 84.95 & 67.74 & 78.71 \\ %\hline
NewsQA &27.99 &53.05 &77.27 &90.41 &89.96 &91.65 & 73.08& 85.88  & 78.85 & 84.30 &69.43 & 81.06\\ %\hline
SQuAD &34.35 & 61.86 &  76.90& 90.51 & 89.94 & 91.07 &  {\bf 77.14} & 85.80  & 78.21 &  {\bf 84.97} &71.31 & 82.84\\ %\hline
QA-RE &35.50 & 65.85 &{\bf 78.30} & {\bf 90.78}& {\bf 90.64}& 91.73&63.36 &85.80  & {\bf 78.94}  & 84.68 &69.35 & 83.77\\ %\hline
Large QA-SRL &{\bf 50.16} & {\bf 72.59} &76.92 & 90.68&90.12 &91.73 &68.99 &85.46 &  78.88 & 84.61 & {\bf 73.01} & {\bf 85.01} \\ %\hline
QAMR &  46.42 &  70.13 &  77.53 &  90.57 &  89.90 & {\bf 92.16} & 72.23&  86.37  &  78.73 & 84.79 & 72.96 & 84.80 \\ \Xhline{2\arrayrulewidth}
\end{tabular}}
\caption{
%\qn{same size everything} Pre-training QASEs on different QA datasets.  We use the same experimental settings as Sec.~\ref{subsec:downstream-tasks} for SDP, NER, RE and TE. For SRL, the strong baseline on OntoNotes is too time-consuming, we instead use a simple BiLSTM model on Propbank as Sec.~\ref{subsec:sample-efficiency}. \hangfeng{add BERT baseline.}
% \qn{I'm thinking of having an extra row that takes the maximum of all those rows below BERT. It tells people that ``using \name{} can improve BERT.''} \dr{good idea!}
Further pre-training \name{} on different QA datasets of the same number of QA pairs ($51K$).
%down sampled if any dataset is larger). 
As we propose, s-\name{} is used as features for single-sentence tasks, and p-\name{} is further fine-tuned for the paired-sentence task.
The specific models are all strong baselines except for SRL, where we use a simple BiLSTM model to save run-time.
``Small'' means 10\% training examples for all tasks except NER, where ``small'' means the dev set (about 23\%) of the corresponding training set. We further show the results of \name{} with the best QA dataset, which are significantly better than those of \bert{}.
% Comparison between \name{} pre-trained on six different QA datasets with the same number of QA pairs ($51K$, the smallest size of the training data in the six QA datasets) and BERT on four single-sentence tasks and one paired-sentence task in the small data setting and the full data setting (full training examples). We use strong baselines for all single-sentence tasks except SRL, where we use a simple BiLSTM model because the strong baseline for SRL is too time-consuming. For all tasks except the NER, we use $10\%$ of the training examples as a small training set. We use the development set ($23\%$) as a small training set for NER. 
%More details in Appendix \ref{sec:experiment-details}.
}
\label{table:pre-training-data}
\end{table*}

\begin{table*}[t]
\centering
\scalebox{0.84}{
\begin{tabular}{c||c|c|c|c|c||c|c||c}
\Xhline{2\arrayrulewidth}
 & \multicolumn{5}{c||}{Single-Sentence Tasks} & \multicolumn{2}{c||}{Paired-Sentence Tasks} &  \\ \hline
 Small & SRL & SDP & NER & RE & Coref & TE & MRC & Avg \\ \hline
 BERT & 76.65 & 75.49& 88.89 & 71.48 & 62.76 &78.29 & 79.90 & 76.21 \\ %\hline
%  QASE & {\bf 80.60} & {\bf 77.53} & {\bf 89.90} & {\bf 72.23} & {\bf 63.36} & {\bf 78.73} & {\bf 82.96} & {\bf 77.90}\\ \hline
 Proposed (abs. imp.) %Absolute Improvement 
  & +3.95 & +2.04 & +1.01 & +0.75 & +0.60 & +0.44 & +3.06 & +1.69\\ %\hline 
  Proposed (rel. imp.)%Relative Improvement 
  & 16.9\% & 8.3\% & 9.1\% & 2.6\% & 1.6\% & 2.0\% & 15.2\% & 7.1\% \\ %\hline
  \hline
 Full & SRL & SDP & NER & RE & Coref & TE & MRC & Avg \\ \hline
 BERT & 84.54& 90.13& 91.38& 86.33& 69.05 &84.09 & 88.23 & 84.82\\ %\hline
%  QASE & {\bf 84.69} & {\bf 90.57} & {\bf 92.16} & {\bf 86.37} & 68.91 & {\bf 84.79} & {\bf 88.58} & {\bf 85.15} \\ \hline
  Proposed (abs. imp.) %Absolute Improvement 
  & +0.15 & +0.44 & +0.78 & +0.04 & -0.14 & +0.7 & +0.35 & +0.33 \\ %\hline
  Proposed (rel. imp.) %Relative Improvement 
  & 0.9\% & 4.5\% & 9.0\% & 0.3\% & -0.5\% & 4.4\% & 3.0\% & 2.2\% \\ \Xhline{2\arrayrulewidth}
\end{tabular}}
\caption{\quase{QAMR} (almost) universally improves on 5 single-sentence tasks and 2 paired-sentence tasks. Note BERT is close to the state of the art for these tasks. Both absolute improvement (abs. imp.) and relative improvement (rel. imp.; error reduction rate) are reported.
``Small/Full'' refers to the size of training data for each target task.
%(not the size of QAMR).
For SDP, RE, TE, and MRC, ``small'' means 10\% of the training set, while for NER, SRL, and Coref, ``small'' means the development set (about 10\%-30\% compared to each training set). 
%More in Appendix \ref{sec:experiment-details}. 
% in the small-data setting and full-data setting. 
% We show the absolute improvement and relative improvement (error reduction rate) rather than the absolute values of \name{}$_{QAMR}$ on the seven tasks. 
% For SDP, RE, TE, and MRC, we use $10\%$ of the training examples as a small training set. For NER, SRL, and Coref, we use the development set (about the same number as $10\%-30\%$ of training examples) as a small training set. 
% We use strong baselines for all five single-sentence tasks and the dataset for SRL is Ontonotes. 
%\qn{the last column, avg, should use multirow}
%\qn{why is a number for coref bold-faced?}
%\qn{%I feel you can change the two rows of ``QASE'' to be showing the number of improvement instead of the absolute numbers. E.g., ``80.60'' can be ``+3.95'', which is more impressive. 
%In addition, the last column isn't impressive, and it's also not fair because the variance of improvement is very large, so I'm considering removing it actually.}\\
%\qn{I think it's ok to show this table later. Readers won't know how to read these numbers and a wrong understanding plays against us.}
}
\label{table:quase-vs-bert}
\end{table*}
%\qn{here the text should summarize the main points in this section instead of reiterating the motivations, which is already explained in the intro.}

%\qn{please use ``\backslach quase[1]'' for \quase{dataset}}
In this section, we conduct thorough experiments to show that \name{} is a good framework to get supervision from QA data for other tasks. 
We first give an overview of the datasets and models used in these experiments before diving into the details of each experiment.
%We first show the necessity of two representations in Sec.~\ref{subsec:necessity-two-representations}. After that, the sample complexity of \name{} compared to BERT on one single-sentence task (SRL) and one paired-sentence task (MRC) are shown in Sec.~\ref{subsec:sample-complexity}. We then analyze the impact of different QA datasets for \name{} to be further pre-trained on in Sec.~\ref{subsec:pre-training-data}. Finally, we evaluate the effectiveness of \quase{QAMR} on 7 tasks (5 single-sentence tasks and 2 paired-sentence tasks). 

Specifically, we use PropBank \cite{kingsbury2002treebank} (SRL), the dataset from the SemEval'15 shared task \cite{oepen-etal-2015-semeval} with DELPH-IN MRS-Derived Semantic Dependencies target representation (SDP), CoNLL'03 \cite{TjongDe03} (NER), the dataset in SemEval'10 Task 8 \cite{hendrickx2009semeval} (RE), the dataset in the CoNLL'12 shared task \cite{pradhan2012conll} (Coref), MNLI \cite{williams-etal-2018-broad} (TE), and SQuAD 1.0 \cite{rajpurkar-etal-2016-squad} (MRC). 
In Table~\ref{table:quase-vs-bert}, we use CoNLL'12 English subset of OntoNotes 5.0 \cite{pradhan2013towards}, which is larger than PropBank.
% We have two datasets for SRL, PropBank \cite{kingsbury2002treebank} and a larger one, CoNLL'12 English subset of OntoNotes 5.0 \cite{pradhan2013towards}. 
% Unless otherwise noted, we show the performance of SRL on PropBank. 
The performance of TE and MRC is evaluated on the development set.\footnote{For TE, we mean matched examples in MNLI.}

For single-sentence tasks, we use both simple baselines (e.g., BiLSTM and CNN; see Appendix \ref{subsec:simple-models}) and near-state-of-the-art models published in recent years.
% As for models in single-sentence tasks, we usually consider two types: simple baselines (see Appendix \ref{subsec:simple-models}) and strong baselines. 
As in ELMo, we use the deep neural model in \citet{he2017deep} for SRL, the model in \citet{peters-etal-2018-deep} for NER, and the end-to-end neural model in \citet{lee-etal-2017-end} for Coref.
We also use the biaffine network in \citet{dozat-manning-2018-simpler} for SDP but we removed part-of-speech tags from its input, and the attention-based BiLSTM in \citet{zhou2016attention} is the strong baseline for RE.
% We use the same strong baselines as ELMo for SRL, NER and Coref: the deep neural model in \cite{he2017deep} for SRL, the model in \cite{peters-etal-2018-deep} for NER, and the end-to-end neural model in \cite{lee-etal-2017-end} for Coref. The biaffine network in \cite{dozat-manning-2018-simpler} without part-of-speech (PoS) tags serves as a strong baseline for SDP, and the attention-based BiLSTM in \cite{zhou2016attention} is the strong baseline for RE. 
In addition, we replace the original word embeddings in these models (e.g., GloVe \cite{PenningtonSoMa14}) by BERT. 
Throughout this paper, we use the pre-trained case-insensitive BERT-base implementation.
More details on our experimental setting
%experiment settings here 
can be found in Appendix~\ref{sec:experiment-details}, including the details of simple models in \ref{subsec:simple-models}, some common experimental settings of \name{} in \ref{subsec:main-experimental-settings}, and s-\name{} combined with other SOTA embeddings (ELMo  and Flair \cite{akbik2018contextual}) in \ref{subsec:other-embeddings}.

\subsection{Necessity of Two Representations}
\label{subsec:necessity-two-representations}
We first consider a straightforward method to use QA data for other tasks---to further pre-train BERT on these QA data. We compare BERT further pre-trained on QAMR (denoted by \bert{QAMR}) with BERT on two single-sentence tasks (SRL and RE) and two paired-sentence tasks (TE and MRC). 
We use
% BERT and BERT (QAMR) 
a feature-based approach for single-sentence tasks and a fine-tuning approach for paired-sentence tasks. The reason is two-fold. On the one hand, current SOTAs of all single-sentence tasks considered in this paper are still feature-based. How to efficiently use sentence encoders (e.g. BERT) in a fine-tuning approach for some complicated tasks (e.g. SRL and SDP) is unclear. On the other hand, the fine-tuning approach shows great advantage over feature-based on many paired-sentence tasks (e.g. TE and MRC).
% The results are shown in Table \ref{table:simple-additional-training}. 
Similar to \citet{phang2018sentence},
% \qn{note the diff between cite and citet}
we find in Table \ref{table:simple-additional-training} that the two single-sentence tasks benefit less than the two paired-sentence tasks from \bert{QAMR}, which indicates that {simply ``further pre-training BERT'' is not enough}.

We then compare s-\quase{QAMR} and p-\quase{QAMR} on three single-sentence tasks (SRL, SDP and NER) and two paired-sentence tasks (TE and MRC) to show that it is important to distinguish two types of sentence representations. 
Rather than concatenating two embeddings as proposed in Sec.~\ref{subsec:two-quase}, 
here we replace BERT embeddings with \name{} embeddings for convenience.
The results are shown in Table \ref{table:s-QASE-vs-p-QASE}. We find that s-\name{} has a great advantage over p-\name{} on single-sentence tasks and p-\name{} is better than s-\name{} on paired-sentence tasks. The proposal of two types of sentence encoders tackles the problem one may encounter when there is only further pre-training BERT on QAMR for single-sentence tasks. In summary, {it is necessary to distinguish two types of sentence representations} for single-sentence tasks and paired-sentence tasks.

\subsection{Sample Complexity of \name{}}
\label{subsec:sample-complexity}
To see whether adding \name{} to BERT reduces the sample complexity, we compare \quase{QAMR} with BERT on one single-sentence task (SRL) and one paired-sentence task (MRC) with different percentages of training examples. For convenience, we replace BERT embeddings with \name{} embeddings for SRL. As shown in Figure \ref{fig:sample-complexity}, we find that s-\quase{QAMR} outperforms BERT on SRL with small training data, and p-\quase{QAMR} outperforms BERT on MRC with small training data. The results support that (1) adding \name{} to BERT reduces the sample complexity, (2) \name{} is very important in the low-resource setting. For instance, s-\quase{QAMR} achieves an F1 score of $61$ in SRL with $30\%$ ($27K$) training examples (compared to $50.92$ F1 by BERT). And p-\quase{QAMR} achieves $69.81$ average F1 on MRC with $0.1\%$ (about $100$) training examples (compared to $13.29$ F1 by BERT).

%To understand the sample efficiency of QASE, we compare QASE with BERT on two specific tasks, SRL and MRC, with different percentages of training examples. \\
%{\bf s-QASE on SRL.} Because the strong baselines of SRL on OntoNotes 5.0 are too time-consuming, we instead use a simple BiLSTM baseline on PropBank \cite{kingsbury2002treebank} here. For simplicity, we replace BERT embeddings with QASE embeddings rather than concatenate these two embeddings. The results of BERT and s-QASE on different percentages of training examples are shown in Figure \ref{fig:bert-qase-srl}. We can see that s-QASE improves the F1 score more with fewer training examples.
%\\
%{\bf p-QASE on MRC.} We show the performance of BERT and p-QASE on MRC with different percentages of training examples. The results in Figure \ref{fig:bert-qase-mrc} indicate that improvements with p-QASE are larger for smaller training sets.

%As shown in Figure \ref{fig:sample-efficiency}, we can know that QASE is more effective in the lower-resource setting. For instance, s-QASE improves BERT from $51$ to $61$ F1 score of SRL with $30\%$ ($27K$) training examples.

\subsection{Data Choice for Further Pre-training}
\label{subsec:pre-training-data}
We compare BERT with \name{} further pre-trained with the same numbre of QA pairs on 6 different QA datasets (TriviaQA \cite{joshi-etal-2017-triviaqa}, NewsQA \cite{trischler-etal-2017-newsqa}, SQuAD, QA-RE \cite{levy-etal-2017-zero}, Large QA-SRL \cite{fitzgerald-etal-2018-large}, and QAMR). s-\name{} further pre-trained on different QA datasets are evaluated on four single-sentence tasks in a feature-based approach: SRL, SDP, NER and RE. p-\name{} further pre-trained on different QA datasets is evaluated on one task (TE) in a fine-tuning approach. 

In Table \ref{table:pre-training-data}, we find that the best options are quite different across different target tasks, which is expected because a task usually benefits more from a more similar QA dataset. 
%\qn{should you push the following paragraph to the next subsection?} 
However, we also find that QAMR is generally a good further-pre-training choice for \name{}. This is consistent with our intuition: First, QAMR has a simpler concept class than other paragraph-level QA datasets, such as TriviaQA, NewsQA and SQuAD. It is easier for \name{} to learn a good representation with QAMR to help sentence-level tasks. Second, QAMR is more general than other sentence-level QA datasets, such as QA-RE and Large QA-SRL.\footnote{Although the average performance of \quase{QAMR} on five tasks is slightly below \quase{Large\;QA-SRL}, for which the benefit mostly comes from SRL. \name{} is mainly designed to improve a lot of tasks, so QAMR is a better choice in our setup, but in practice, we do not limit \name{} to any specific QA dataset and one can use the best one for corresponding target tasks.}
% we can still say that QAMR is more general than Large QA-SRL. Because \quase{Large\;QA-SRL} has much better performance on SRL, where Large QA-SRL is too related to SRL. }, 
Therefore, we think that the capability to identify predicate-argument structures can generally help many sentence-level tasks, as we discuss next.
% so \name{} further pre-trained on QAMR can help more sentence-level tasks.
\subsection{The Effectiveness of \name{}}
\label{subsec:effectiveness-quase}
%\qn{Should we call this subsection ``dataset''?}

%{\bf Single-Sentence Tasks.}  We use the CoNLL-2012 English subset of OntoNotes 5.0 \cite{pradhan2013towards} and the deep neural model in \cite{he2017deep} for SRL. We use the dataset from SemEval 2015 shared task \cite{oepen-etal-2015-semeval} with DELPH-IN MRS-Derived Semantic Dependencies (DM) target representation and the biaffine network in \cite{dozat-manning-2018-simpler} without part-of-speech (PoS) tags for SDP. We use CoNLL 2003 dataset \cite{TjongDe03} and the model in \cite{peters-etal-2018-deep} for NER. We use the dataset in Semeval 2010 Task 8 \cite{hendrickx2009semeval} and the attention-based BiLSTM in \cite{zhou2016attention} for RE. We use the dataset in CoNLL 2012 shared task \cite{pradhan2012conll} and the end-to-end neural model in \cite{lee-etal-2017-end} for Coref.
%\\
%{\bf Paired-Sentence Tasks.} We use SQuAD 1.0 \cite{rajpurkar-etal-2016-squad} as our dataset for MRC, and MNLI \cite{williams-etal-2018-broad} as our dataset for TE. For both tasks, fine-tuning based BERT serves as our baseline and the performance is evaluated on the development set (we use the development set for matched examples for MNLI).
Here we compare \quase{QAMR} with BERT on 5 single-sentence tasks and 2 paired-sentence tasks, where \quase{QAMR} is further pre-trained on the training set ($51K$ QA pairs) of the QAMR dataset. As shown in Table \ref{table:quase-vs-bert}, we find that \quase{QAMR} has a better performance than BERT on both single-sentence tasks and paired-sentence tasks, especially in the low-resource setting\footnote{Another interesting finding is that simple models usually benefit more from \name{} embeddings than SOTA models.}, indicating that \quase{QAMR} can provide extra features compared to BERT. 

Admittedly, the improvement in the ``Full'' setting is not significantly large, but we think that this is expected because large direct training data are available (such as SRL with $278K$ training examples in OntoNotes). 
% that further pre-training with $51K$ indirect QA pairs cannot improve downstream tasks by a large margin when large direct training data are available (such as SRL with $278K$ training examples in OntoNotes). 
However, it is still promising that $51K$ indirect QA pairs can improve downstream tasks in the low-resource setting (i.e. several thousands direct training examples). That is because they help the scalability of machine learning methods, especially for some specific domains or some low-resource languages where direct training data do not exist in large scale.

\section{Discussion}
\label{sec:discussions}
In this section we discuss a few issues pertaining to improving \name{} by using additional QA datasets and the comparison of \name{} with related symbolic representations.
\subsection{Further Pre-training \name{} on Multiple QA Datasets}
\label{subsec:existing-resource}
%\qn{really it's not ``existing resource'' (which isn't informational at all); it's ``multiple QA data'' at one time, isn't it?}
We investigate whether adding the Large QA-SRL dataset \cite{fitzgerald-etal-2018-large} or the QA-RE\footnote{Because the training set of QA-RE is too large, we randomly choose $100,000$ training examples.} dataset into QAMR in the further pre-training stage can help SRL and RE. We use s-\name{} embeddings to replace BERT embeddings instead of concatenating the two embeddings. The effectiveness of adding existing resources (Large QA-SRL or QA-RE) into QAMR in the further pre-training stage of s-\name{} on SRL and RE are shown in Table \ref{table:add-existing-resource}. We find that adding related QA signals (Large QA-SRL for SRL and QA-RE for RE) into QAMR can help improve specific tasks. Noteworthy is the fact that QA-RE can help SRL (Large QA-SRL can also help RE), though the improvement is minor compared to Large QA-SRL (QA-RE). These results indicate that adding more QA signals related to the sentence can help get a better sentence representation in general.
\begin{table}
\centering
\scalebox{0.72}{
\begin{tabular}{c||c|c||c|c}
\Xhline{2\arrayrulewidth}
 Tasks & \multicolumn{2}{c|}{SRL} & \multicolumn{2}{c}{RE} \\ \hline
Split & 10\% & 100\% & 10\% & 100\% \\ \hline
BERT& 34.16  & 66.02 &59.36 & 83.28\\ \hline
\quase{QAMR} & 46.42  & 70.13  & 61.09& 82.22 \\ \hline
\quase{QAMR+Large\; QA-SRL} & {\bf 49.92}  & 71.74 &65.76 & 83.16\\ \hline
\quase{QAMR+QA-RE} & 47.25  & {\bf 72.52} & {\bf 68.12} & {\bf 83.89} \\ \Xhline{2\arrayrulewidth}
\end{tabular}}
\caption{
The potential of further improving \quase{QAMR} by further pre-training it on more QA data. The ``+'' between datasets means union with shuffling. Both Large QA-SRL and QA-RE help achieve better results than QAMR alone.
% The effectiveness of adding existing resources, Large QA-SRL and QA-RE into QAMR, in the pre-training stage of s-\name{} on SRL and RE. 
For simplicity, we use a simple BiLSTM model for SRL and a simple CNN model for RE. See more in Appendix \ref{sec:experiment-details}.
}
\label{table:add-existing-resource}
\end{table} 
\subsection{Comparison with Symbolic Meaning Representations}
\label{subsec:symbolic-semantics}
%\dr{I think that the next sentence conflates two issues. I modified it}
Traditional (symbolic) shallow meaning representations such as SRL and AMR, suffer from having a fixed set of relations one has to commit to. Moreover, inducing these representations requires costly annotation by experts.
%Traditional symbolic meaning representations, such as SRL and AMR, suffer from costly annotations and are not flexible because of the pre-defined formalism. 
%
Proposals such as QA-SRL, QAMR, semantic proto-roles \cite{reisinger2015semantic}, and universal dependencies \cite{universal-decompositional-semantics-on-universal-dependencies}
%\dr{{\bf Add here Ben Van Durme's formalisms}; we must cite his work here} 
avoid some of these issues by using natural language annotations, but it is unclear how other tasks can take advantage of them. \name{} is proposed to facilitate inducing distributed representations instead of symbolic representations from QA signals; it benefits from cheaper annotation and flexibility, and can also be easily used in downstream tasks.
%For example, given the sentence "Dan gave a book to Jim's brother, Joe.", p-\name{} can correctly answer "Who is Jim's brother?" and "Who is Joe's brother?". It is not surprising that p-\name{} can answer the first question, because it is also covered by AMR. However, the symmetry of this relation is not covered by AMR, but we can get it through QA signals.

The following probing analysis, based on the Xinhua subset in the AMR dataset\drc{,} shows that s-\quase{QAMR} embeddings encode more semantics related to AMR than BERT embeddings. Specifically, we use the same edge probing model as \citet{tenney2019you}, and find that the probing accuracy ($73.59$) of s-\quase{QAMR} embeddings is higher than that ($71.58$) of BERT. At the same time, we find that p-\quase{QAMR} can achieve $76.91$ F1 on the PTB set
of QA-SRL, indicating that p-\quase{QAMR} can capture enough information related to SRL to have a good zero-shot SRL performance. More details can be found in Appendix \ref{subsec:symbol-representations-details}. Another fact worth noting is that AMR can be used to improve downstream tasks, such as MRC \cite{sachan2016machine}, TE \cite{lien2015semantic}, RE \cite{garg2019kernelized} and SRL \cite{song2018easier}. The benefits of \quase{QAMR} on downstream tasks show that we can take advantage of AMR by learning from much cheaper QA signals dedicated to it.
\subsection{Difficulties in Learning Symbolic Representations from QA Signals}
\label{subsec:symbol-QA}
\name{} is designed to learn distributed representations from QA signals to help down-stream tasks. We further show 
%A natural question is: Can we learn symbolic representations instead of distributed representations from QA signals? We answer it by analyzing 
the difficulties of learning two types of corresponding symbolic representations from QA signals, which indicates that the two other possible methods are not as tractable as ours. 

%{\bf Learning QAMR Representations from QA Signals.} 
One option of symbolic representation is the QAMR graph. \citet{MSHDZ17} show that question generation for QAMR representations can only achieve a precision of $28\%$, and a recall of $24\%$, even with fuzzy matching (multi-BLEU\footnote{An average of BLEU1--BLEU4 scores.} $>0.8$). Furthermore, it is still unclear how to use the complex QAMR graph in downstream tasks. \textit{These results indicate that learning a QAMR parser for down-stream tasks is mainly hindered by question generation,  and  how  to  use  the  full  information of QAMR for downstream tasks is still unclear.}  
%{\bf Learning SRL/AMR Representations from QA Signals.}

Another choice of symbolic representation is AMR, since QAMR is proposed to replace AMR. We consider a simpler setting, learning an SRL parser from Large QA-SRL. 
%We first design a reasonable upperbound by treating the answers that have overlapped with some arguments as our predicted arguments. After that, 
We propose three models in different perspectives, but the best performance of them is only $54.10$ F1, even with fuzzy matching (Intersection/Union $\geq$ $0.5$). More details can be found in Appendix \ref{subsec:srl-qa-details}. Although a lot of methods \cite{KKSR18, marcheggiani-titov-2017-encoding, strubell-etal-2018-linguistically} can be adopted to use SRL/AMR in downstream tasks, \textit{the difficulty of learning a good SRL/AMR parser from QA signals hinders
%blocks 
this direction.}

The difficulties of learning the two types of symbolic representations from QA signals indicate that our proposal of learning distributed representations from QA signals is a better way of making
%good choice to make 
use of the latent semantic information in QA pairs for down-stream tasks.
\section{Conclusion}
\label{sec:conclusion}
In this paper, we investigate an important problem in NLP: Can we make use of low-cost signals, such as QA data, to help related tasks? We retrieve signals from sentence-level QA pairs to help NLP tasks via two types of sentence encoding approaches. For tasks with a single-sentence input, such as SRL and NER, we propose s-\name{} that provides latent sentence-level representations; 
for tasks with a sentence pair input, such as TE and MRC we propose p-\name{}, that generates latent representations related to attentions. Experiments on a wide range of tasks show that the distinction of s-\name{} and p-\name{} is highly 
%especially 
effective, and \quase{QAMR} has the potential to improve on many tasks,
%in general, 
especially in the low-resource setting.  
%\hfch{We plan to improve \name{} with the help of stronger language models, such as XLNet or RoBERTa, as our future work.}

% Future work involves two directions. First, it will be interesting to see how existing resources can be utilized by \name{}. An example is to design some heuristic rules to generate simple QA pairs from a Coref dataset. Second, we plan to improve \name{} with the help of stronger language models, such as XLNet or RoBERTa. 
% \qn{I find it not super exciting to mention these two future directions.} \dr{do we want to give the first idea away?}

%\qn{remember to submit the appendix as a separate file; submitting it with the main pdf will lead to rejection.}

\section*{Acknowledgements}

This material is based upon work supported by the US Defense Advanced Research Projects Agency (DARPA) under contracts FA8750-19-2-0201,  W911NF-15-1-0461, and FA8750-19-2-1004, a grant from the Army Research Office (ARO), and Google Cloud. This research is also based upon work supported in part by the Oﬃce of the Director of National Intelligence (ODNI), Intelligence Advanced Research Projects Activity (IARPA), via IARPA Contract No. 2019-19051600006 under the BETTER Program. The views expressed are those of the authors and do not reflect the official policy or position of the Department of Defense or the U.S. Government.

%\dr{Note that in the reference for Sentence-BERT (and maybe in others), you need to put{BERT} in curly brackets so it will be kept capitalized}

\bibliography{cited-long,new}

\begin{thebibliography}{52}
\expandafter\ifx\csname natexlab\endcsname\relax\def\natexlab#1{#1}\fi

\bibitem[{Akbik et~al.(2018)Akbik, Blythe, and Vollgraf}]{akbik2018contextual}
Alan Akbik, Duncan Blythe, and Roland Vollgraf. 2018.
\newblock Contextual string embeddings for sequence labeling.
\newblock In \emph{COLING}, pages 1638--1649.

\bibitem[{Chang et~al.(2007)Chang, Ratinov, and Roth}]{ChangRaRo07}
M.~Chang, L.~Ratinov, and D.~Roth. 2007.
\newblock Guiding semi-supervision with constraint-driven learning.
\newblock In \emph{Proc. of the Annual Meeting of the Association for
  Computational Linguistics (ACL)}, pages 280--287.

\bibitem[{Devlin et~al.(2019)Devlin, Chang, Lee, and
  Toutanova}]{devlin-etal-2019-bert}
Jacob Devlin, Ming-Wei Chang, Kenton Lee, and Kristina Toutanova. 2019.
\newblock {BERT}: Pre-training of deep bidirectional transformers for language
  understanding.
\newblock In \emph{NAACL}, pages 4171--4186.

\bibitem[{Dozat and Manning(2018)}]{dozat-manning-2018-simpler}
Timothy Dozat and Christopher~D. Manning. 2018.
\newblock Simpler but more accurate semantic dependency parsing.
\newblock In \emph{ACL}, pages 484--490.

\bibitem[{FitzGerald et~al.(2018)FitzGerald, Michael, He, and
  Zettlemoyer}]{fitzgerald-etal-2018-large}
Nicholas FitzGerald, Julian Michael, Luheng He, and Luke Zettlemoyer. 2018.
\newblock Large-scale {QA}-{SRL} parsing.
\newblock In \emph{ACL}, pages 2051--2060.

\bibitem[{Gardner et~al.(2017)Gardner, Grus, Neumann, Tafjord, Dasigi, Liu,
  Peters, Schmitz, and Zettlemoyer}]{Gardner2017AllenNLP}
Matt Gardner, Joel Grus, Mark Neumann, Oyvind Tafjord, Pradeep Dasigi,
  Nelson~F. Liu, Matthew Peters, Michael Schmitz, and Luke~S. Zettlemoyer.
  2017.
\newblock \href {http://arxiv.org/abs/arXiv:1803.07640} {{AllenNLP}: A deep
  semantic natural language processing platform}.

\bibitem[{Garg et~al.(2019)Garg, Galstyan, Ver~Steeg, Rish, Cecchi, and
  Gao}]{garg2019kernelized}
Sahil Garg, Aram Galstyan, Greg Ver~Steeg, Irina Rish, Guillermo Cecchi, and
  Shuyang Gao. 2019.
\newblock Kernelized hashcode representations for relation extraction.
\newblock In \emph{Proceedings of the AAAI Conference on Artificial
  Intelligence}, volume~33, pages 6431--6440.

\bibitem[{He et~al.(2017)He, Lee, Lewis, and Zettlemoyer}]{he2017deep}
Luheng He, Kenton Lee, Mike Lewis, and Luke Zettlemoyer. 2017.
\newblock Deep semantic role labeling: What works and what's next.
\newblock In \emph{ACL}, pages 473--483.

\bibitem[{He et~al.(2015)He, Lewis, and Zettlemoyer}]{HeLeZe15}
Luheng He, Mike Lewis, and Luke Zettlemoyer. 2015.
\newblock Question-answer driven semantic role labeling: Using natural language
  to annotate natural language.
\newblock In \emph{Proc. of the Conference on Empirical Methods for Natural
  Language Processing (EMNLP)}, pages 643--653.

\bibitem[{Hendrickx et~al.(2009)Hendrickx, Kim, Kozareva, Nakov,
  {\'O}~S{\'e}aghdha, Pad{\'o}, Pennacchiotti, Romano, and
  Szpakowicz}]{hendrickx2009semeval}
Iris Hendrickx, Su~Nam Kim, Zornitsa Kozareva, Preslav Nakov, Diarmuid
  {\'O}~S{\'e}aghdha, Sebastian Pad{\'o}, Marco Pennacchiotti, Lorenza Romano,
  and Stan Szpakowicz. 2009.
\newblock Semeval-2010 task 8: Multi-way classification of semantic relations
  between pairs of nominals.
\newblock In \emph{Proceedings of the Workshop on Semantic Evaluations: Recent
  Achievements and Future Directions}, pages 94--99.

\bibitem[{Joshi et~al.(2017)Joshi, Choi, Weld, and
  Zettlemoyer}]{joshi-etal-2017-triviaqa}
Mandar Joshi, Eunsol Choi, Daniel Weld, and Luke Zettlemoyer. 2017.
\newblock {T}rivia{QA}: A large scale distantly supervised challenge dataset
  for reading comprehension.
\newblock In \emph{Proceedings of the 55th Annual Meeting of the Association
  for Computational Linguistics (Volume 1: Long Papers)}, pages 1601--1611.
  Association for Computational Linguistics.

\bibitem[{Khashabi et~al.(2018)Khashabi, Khot, Sabharwal, and Roth}]{KKSR18}
Daniel Khashabi, Tushar Khot, Ashish Sabharwal, and Dan Roth. 2018.
\newblock Question answering as global reasoning over semantic abstractions.
\newblock In \emph{Proceedings of The Conference on Artificial Intelligence
  (Proc. of the Conference on Artificial Intelligence (AAAI))}.

\bibitem[{Kingsbury and Palmer(2002)}]{kingsbury2002treebank}
Paul Kingsbury and Martha Palmer. 2002.
\newblock From treebank to propbank.
\newblock In \emph{LREC}, pages 1989--1993. Citeseer.

\bibitem[{Kiros et~al.(2015)Kiros, Zhu, Salakhutdinov, Zemel, Urtasun,
  Torralba, and Fidler}]{kiros2015skip}
Ryan Kiros, Yukun Zhu, Russ~R Salakhutdinov, Richard Zemel, Raquel Urtasun,
  Antonio Torralba, and Sanja Fidler. 2015.
\newblock Skip-thought vectors.
\newblock In \emph{Advances in neural information processing systems}, pages
  3294--3302.

\bibitem[{Lee et~al.(2017)Lee, He, Lewis, and Zettlemoyer}]{lee-etal-2017-end}
Kenton Lee, Luheng He, Mike Lewis, and Luke Zettlemoyer. 2017.
\newblock End-to-end neural coreference resolution.
\newblock In \emph{EMNLP}, pages 188--197.

\bibitem[{Levy et~al.(2017)Levy, Seo, Choi, and
  Zettlemoyer}]{levy-etal-2017-zero}
Omer Levy, Minjoon Seo, Eunsol Choi, and Luke Zettlemoyer. 2017.
\newblock Zero-shot relation extraction via reading comprehension.
\newblock In \emph{CoNLL}, pages 333--342.

\bibitem[{Lien and Kouylekov(2015)}]{lien2015semantic}
Elisabeth Lien and Milen Kouylekov. 2015.
\newblock Semantic parsing for textual entailment.
\newblock In \emph{Proceedings of the 14th International Conference on Parsing
  Technologies}, pages 40--49.

\bibitem[{Liu et~al.(2019)Liu, Ott, Goyal, Du, Joshi, Chen, Levy, Lewis,
  Zettlemoyer, and Stoyanov}]{liu2019roberta}
Yinhan Liu, Myle Ott, Naman Goyal, Jingfei Du, Mandar Joshi, Danqi Chen, Omer
  Levy, Mike Lewis, Luke Zettlemoyer, and Veselin Stoyanov. 2019.
\newblock Ro{BERT}a: A robustly optimized {BERT} pretraining approach.
\newblock \emph{arXiv preprint arXiv:1907.11692}.

\bibitem[{Marcheggiani and Titov(2017)}]{marcheggiani-titov-2017-encoding}
Diego Marcheggiani and Ivan Titov. 2017.
\newblock Encoding sentences with graph convolutional networks for semantic
  role labeling.
\newblock In \emph{EMNLP}, pages 1506--1515.

\bibitem[{McCann et~al.(2017)McCann, Bradbury, Xiong, and
  Socher}]{mccann2017learned}
Bryan McCann, James Bradbury, Caiming Xiong, and Richard Socher. 2017.
\newblock Learned in translation: Contextualized word vectors.
\newblock In \emph{NeurIPS}, pages 6294--6305.

\bibitem[{Michael et~al.(2017)Michael, Stanovsky, He, Dagan, and
  Zettlemoyer}]{MSHDZ17}
Julian Michael, Gabriel Stanovsky, Luheng He, Ido Dagan, and Luke Zettlemoyer.
  2017.
\newblock Crowdsourcing question-answer meaning representations.
\newblock \emph{NAACL}.

\bibitem[{Oepen et~al.(2015)Oepen, Kuhlmann, Miyao, Zeman, Cinkov{\'a},
  Flickinger, Haji{\v{c}}, and Ure{\v{s}}ov{\'a}}]{oepen-etal-2015-semeval}
Stephan Oepen, Marco Kuhlmann, Yusuke Miyao, Daniel Zeman, Silvie Cinkov{\'a},
  Dan Flickinger, Jan Haji{\v{c}}, and Zde{\v{n}}ka Ure{\v{s}}ov{\'a}. 2015.
\newblock {S}em{E}val 2015 task 18: Broad-coverage semantic dependency parsing.
\newblock In \emph{Proceedings of the 9th International Workshop on Semantic
  Evaluation ({S}em{E}val 2015)}, pages 915--926.

\bibitem[{Palmer et~al.(2010)Palmer, Gildea, and Xue}]{PalmerGiXu10}
M.~Palmer, D.~Gildea, and N.~Xue. 2010.
\newblock \emph{Semantic Role Labeling}.

\bibitem[{Pennington et~al.(2014)Pennington, Socher, and
  Manning}]{PenningtonSoMa14}
Jeffrey Pennington, Richard Socher, and Christopher~D. Manning. 2014.
\newblock Glove: Global vectors for word representation.
\newblock In \emph{EMNLP}, pages 1532--1543.

\bibitem[{Peters et~al.(2018)Peters, Neumann, Iyyer, Gardner, Clark, Lee, and
  Zettlemoyer}]{peters-etal-2018-deep}
Matthew Peters, Mark Neumann, Mohit Iyyer, Matt Gardner, Christopher Clark,
  Kenton Lee, and Luke Zettlemoyer. 2018.
\newblock Deep contextualized word representations.
\newblock In \emph{NAACL}, pages 2227--2237.

\bibitem[{Phang et~al.(2018)Phang, F{\'e}vry, and Bowman}]{phang2018sentence}
Jason Phang, Thibault F{\'e}vry, and Samuel~R Bowman. 2018.
\newblock Sentence encoders on stilts: Supplementary training on intermediate
  labeled-data tasks.
\newblock \emph{arXiv preprint arXiv:1811.01088}.

\bibitem[{Pradhan et~al.(2013)Pradhan, Moschitti, Xue, Ng, Bj{\"o}rkelund,
  Uryupina, Zhang, and Zhong}]{pradhan2013towards}
Sameer Pradhan, Alessandro Moschitti, Nianwen Xue, Hwee~Tou Ng, Anders
  Bj{\"o}rkelund, Olga Uryupina, Yuchen Zhang, and Zhi Zhong. 2013.
\newblock Towards robust linguistic analysis using ontonotes.
\newblock In \emph{CoNLL}, pages 143--152.

\bibitem[{Pradhan et~al.(2012)Pradhan, Moschitti, Xue, Uryupina, and
  Zhang}]{pradhan2012conll}
Sameer Pradhan, Alessandro Moschitti, Nianwen Xue, Olga Uryupina, and Yuchen
  Zhang. 2012.
\newblock Conll-2012 shared task: Modeling multilingual unrestricted
  coreference in ontonotes.
\newblock In \emph{Joint Conference on EMNLP and CoNLL-Shared Task}, pages
  1--40.

\bibitem[{Radford et~al.(2018)Radford, Narasimhan, Salimans, and
  Sutskever}]{radford2018improving}
Alec Radford, Karthik Narasimhan, Tim Salimans, and Ilya Sutskever. 2018.
\newblock Improving language understanding by generative pre-training.
\newblock \emph{URL https://s3-us-west-2. amazonaws.
  com/openai-assets/researchcovers/languageunsupervised/language understanding
  paper. pdf}.

\bibitem[{Rajpurkar et~al.(2016)Rajpurkar, Zhang, Lopyrev, and
  Liang}]{rajpurkar-etal-2016-squad}
Pranav Rajpurkar, Jian Zhang, Konstantin Lopyrev, and Percy Liang. 2016.
\newblock {SQ}u{AD}: 100,000+ questions for machine comprehension of text.
\newblock In \emph{EMNLP}, pages 2383--2392.

\bibitem[{Reimers and Gurevych(2019)}]{reimers2019sentence}
Nils Reimers and Iryna Gurevych. 2019.
\newblock Sentence-{BERT}: Sentence embeddings using siamese {BERT}-networks.
\newblock In \emph{Proceedings of the 2019 Conference on Empirical Methods in
  Natural Language Processing and the 9th International Joint Conference on
  Natural Language Processing (EMNLP-IJCNLP)}, pages 3973--3983.

\bibitem[{Reisinger et~al.(2015)Reisinger, Rudinger, Ferraro, Harman, Rawlins,
  and Van~Durme}]{reisinger2015semantic}
Drew Reisinger, Rachel Rudinger, Francis Ferraro, Craig Harman, Kyle Rawlins,
  and Benjamin Van~Durme. 2015.
\newblock Semantic proto-roles.
\newblock \emph{Transactions of the Association for Computational Linguistics},
  3:475--488.

\bibitem[{Roth(2017)}]{Roth17}
Dan Roth. 2017.
\newblock Incidental supervision: Moving beyond supervised learning.
\newblock In \emph{Proc. of the Conference on Artificial Intelligence (AAAI)}.

\bibitem[{Sachan and Xing(2016)}]{sachan2016machine}
Mrinmaya Sachan and Eric Xing. 2016.
\newblock Machine comprehension using rich semantic representations.
\newblock In \emph{Proceedings of the 54th Annual Meeting of the Association
  for Computational Linguistics (Volume 2: Short Papers)}, pages 486--492.

\bibitem[{Seo et~al.(2017)Seo, Kembhavi, Farhadi, and
  Hajishirzi}]{seo2016bidirectional}
Minjoon Seo, Aniruddha Kembhavi, Ali Farhadi, and Hannaneh Hajishirzi. 2017.
\newblock Bidirectional attention flow for machine comprehension.
\newblock \emph{ICLR}.

\bibitem[{Song et~al.(2018)Song, Wen, Ge, Li, Zhou, Qu, and
  Xue}]{song2018easier}
Li~Song, Yuan Wen, Sijia Ge, Bin Li, Junsheng Zhou, Weiguang Qu, and Nianwen
  Xue. 2018.
\newblock An easier and efficient framework to annotate semantic roles:
  Evidence from the chinese amr corpus.
\newblock In \emph{The 13th Workshop on Asian Language Resources}, page~29.

\bibitem[{Strubell et~al.(2018)Strubell, Verga, Andor, Weiss, and
  McCallum}]{strubell-etal-2018-linguistically}
Emma Strubell, Patrick Verga, Daniel Andor, David Weiss, and Andrew McCallum.
  2018.
\newblock Linguistically-informed self-attention for semantic role labeling.
\newblock In \emph{EMNLP}, pages 5027--5038.

\bibitem[{Sun et~al.(2019)Sun, Yu, Yu, and Cardie}]{sun-etal-2019-improving}
Kai Sun, Dian Yu, Dong Yu, and Claire Cardie. 2019.
\newblock Improving machine reading comprehension with general reading
  strategies.
\newblock In \emph{Proceedings of the 2019 Conference of the North {A}merican
  Chapter of the Association for Computational Linguistics: Human Language
  Technologies, Volume 1 (Long and Short Papers)}, pages 2633--2643,
  Minneapolis, Minnesota. Association for Computational Linguistics.

\bibitem[{Talmor and Berant(2019)}]{talmor-berant-2019-multiqa}
Alon Talmor and Jonathan Berant. 2019.
\newblock {M}ulti{QA}: An empirical investigation of generalization and
  transfer in reading comprehension.
\newblock In \emph{Proceedings of the 57th Annual Meeting of the Association
  for Computational Linguistics}, pages 4911--4921. Association for
  Computational Linguistics.

\bibitem[{Tenney et~al.(2019)Tenney, Xia, Chen, Wang, Poliak, McCoy, Kim,
  Van~Durme, Bowman, Das et~al.}]{tenney2019you}
Ian Tenney, Patrick Xia, Berlin Chen, Alex Wang, Adam Poliak, R~Thomas McCoy,
  Najoung Kim, Benjamin Van~Durme, Samuel~R Bowman, Dipanjan Das, et~al. 2019.
\newblock What do you learn from context? probing for sentence structure in
  contextualized word representations.
\newblock \emph{ICLR}.

\bibitem[{Tishby et~al.(1999)Tishby, Pereira, and
  Bialek}]{tishby2000information}
Naftali Tishby, Fernando~C Pereira, and William Bialek. 1999.
\newblock The information bottleneck method.
\newblock In \emph{Proc. of the Annual Allerton Conference on Communication,
  Control and Computing}.

\bibitem[{Tishby and Zaslavsky(2015)}]{tishby2015deep}
Naftali Tishby and Noga Zaslavsky. 2015.
\newblock Deep learning and the information bottleneck principle.
\newblock In \emph{IEEE Information Theory Workshop (ITW)}.

\bibitem[{Tjong Kim~Sang and De~Meulder(2003)}]{TjongDe03}
Erik~F Tjong Kim~Sang and Fien De~Meulder. 2003.
\newblock Introduction to the conll-2003 shared task: Language-independent
  named entity recognition.
\newblock In \emph{Proc. of the Annual Meeting of the North American
  Association of Computational Linguistics (NAACL)}.

\bibitem[{Trischler et~al.(2017)Trischler, Wang, Yuan, Harris, Sordoni,
  Bachman, and Suleman}]{trischler-etal-2017-newsqa}
Adam Trischler, Tong Wang, Xingdi Yuan, Justin Harris, Alessandro Sordoni,
  Philip Bachman, and Kaheer Suleman. 2017.
\newblock {N}ews{QA}: A machine comprehension dataset.
\newblock In \emph{Proceedings of the 2nd Workshop on Representation Learning
  for {NLP}}, pages 191--200. Association for Computational Linguistics.

\bibitem[{Vaswani et~al.(2017)Vaswani, Shazeer, Parmar, Uszkoreit, Jones,
  Gomez, Kaiser, and Polosukhin}]{vaswani2017attention}
Ashish Vaswani, Noam Shazeer, Niki Parmar, Jakob Uszkoreit, Llion Jones,
  Aidan~N Gomez, {\L}ukasz Kaiser, and Illia Polosukhin. 2017.
\newblock Attention is all you need.
\newblock In \emph{Advances in neural information processing systems}, pages
  5998--6008.

\bibitem[{Wang et~al.(2019)Wang, Hula, Xia, Pappagari, McCoy, Patel, Kim,
  Tenney, Huang, Yu et~al.}]{wang2019can}
Alex Wang, Jan Hula, Patrick Xia, Raghavendra Pappagari, R~Thomas McCoy, Roma
  Patel, Najoung Kim, Ian Tenney, Yinghui Huang, Katherin Yu, et~al. 2019.
\newblock Can you tell me how to get past sesame street? sentence-level
  pretraining beyond language modeling.
\newblock In \emph{Proceedings of the 57th Annual Meeting of the Association
  for Computational Linguistics}, pages 4465--4476.

\bibitem[{Wang et~al.(2018)Wang, Singh, Michael, Hill, Levy, and
  Bowman}]{wang-etal-2018-glue}
Alex Wang, Amanpreet Singh, Julian Michael, Felix Hill, Omer Levy, and Samuel
  Bowman. 2018.
\newblock {GLUE}: A multi-task benchmark and analysis platform for natural
  language understanding.
\newblock In \emph{Proceedings of the 2018 {EMNLP} Workshop {B}lackbox{NLP}:
  Analyzing and Interpreting Neural Networks for {NLP}}, pages 353--355.

\bibitem[{White et~al.(2016)White, Reisinger, Sakaguchi, Vieira, Zhang,
  Rudinger, Rawlins, and {Van
  Durme}}]{universal-decompositional-semantics-on-universal-dependencies}
Aaron~Steven White, Drew Reisinger, Keisuke Sakaguchi, Tim Vieira, Sheng Zhang,
  Rachel Rudinger, Kyle Rawlins, and Benjamin {Van Durme}. 2016.
\newblock {Universal Decompositional Semantics on Universal Dependencies}.
\newblock In \emph{Empirical Methods in Natural Language Processing (EMNLP)}.

\bibitem[{Williams et~al.(2018)Williams, Nangia, and
  Bowman}]{williams-etal-2018-broad}
Adina Williams, Nikita Nangia, and Samuel Bowman. 2018.
\newblock A broad-coverage challenge corpus for sentence understanding through
  inference.
\newblock In \emph{NAACL}, pages 1112--1122.

\bibitem[{Wolf et~al.(2019)Wolf, Debut, Sanh, Chaumond, Delangue, Moi, Cistac,
  Rault, Louf, Funtowicz, and Brew}]{Wolf2019HuggingFacesTS}
Thomas Wolf, Lysandre Debut, Victor Sanh, Julien Chaumond, Clement Delangue,
  Anthony Moi, Pierric Cistac, Tim Rault, R'emi Louf, Morgan Funtowicz, and
  Jamie Brew. 2019.
\newblock Huggingface's transformers: State-of-the-art natural language
  processing.
\newblock \emph{ArXiv}, abs/1910.03771.

\bibitem[{Yang et~al.(2019)Yang, Dai, Yang, Carbonell, Salakhutdinov, and
  Le}]{yang2019xlnet}
Zhilin Yang, Zihang Dai, Yiming Yang, Jaime Carbonell, Ruslan Salakhutdinov,
  and Quoc~V Le. 2019.
\newblock {XLNet}: Generalized autoregressive pretraining for language
  understanding.
\newblock \emph{arXiv preprint arXiv:1906.08237}.

\bibitem[{Zhou et~al.(2016)Zhou, Shi, Tian, Qi, Li, Hao, and
  Xu}]{zhou2016attention}
Peng Zhou, Wei Shi, Jun Tian, Zhenyu Qi, Bingchen Li, Hongwei Hao, and Bo~Xu.
  2016.
\newblock Attention-based bidirectional long short-term memory networks for
  relation classification.
\newblock In \emph{ACL}, pages 207--212.

\end{thebibliography}
\bibliographystyle{acl_natbib}

\clearpage
\appendix
\begin{figure*}[ht]
		\centering
		\subfigure[Error analysis for \name{}.]{
			\centering
			\includegraphics[scale=0.39]{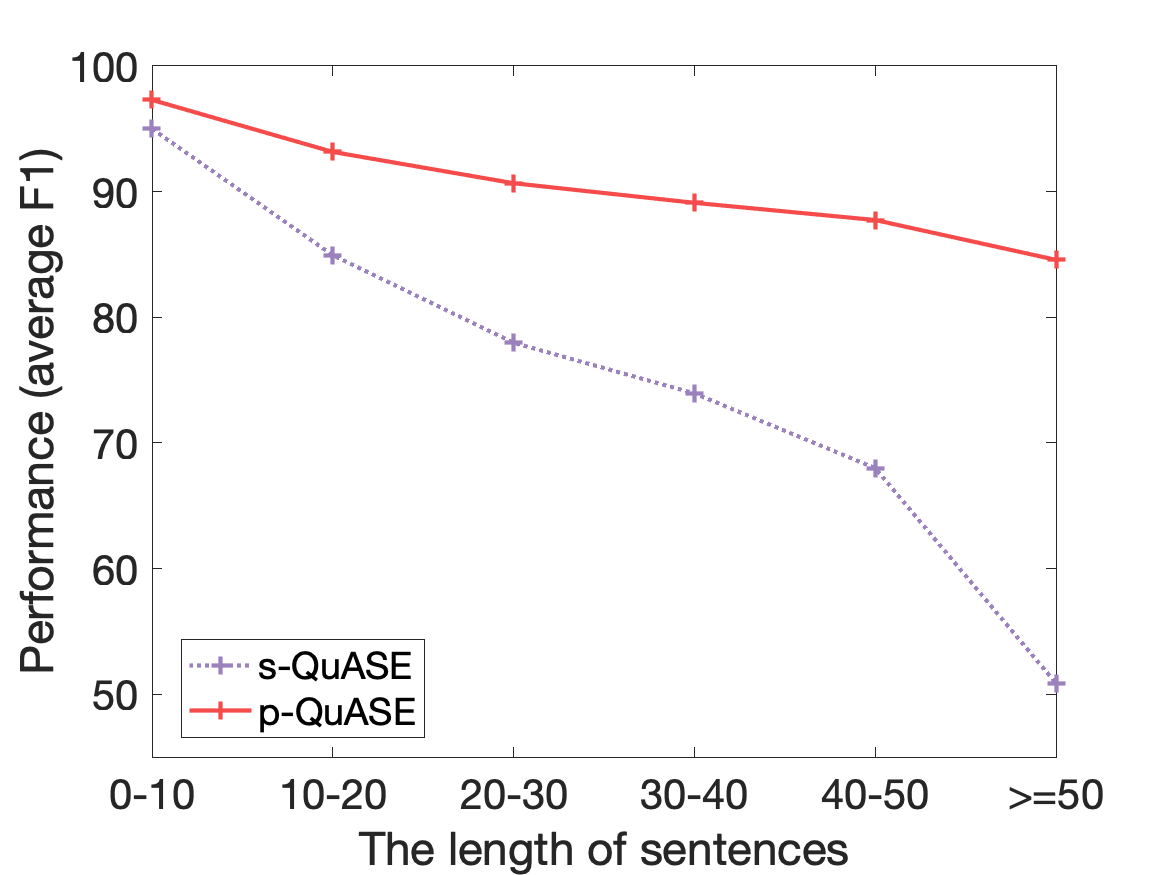}
			\label{fig:error-sent-f1}}
     	\subfigure[Number of QA pairs.]{
     	\centering
     	\includegraphics[scale=0.39]{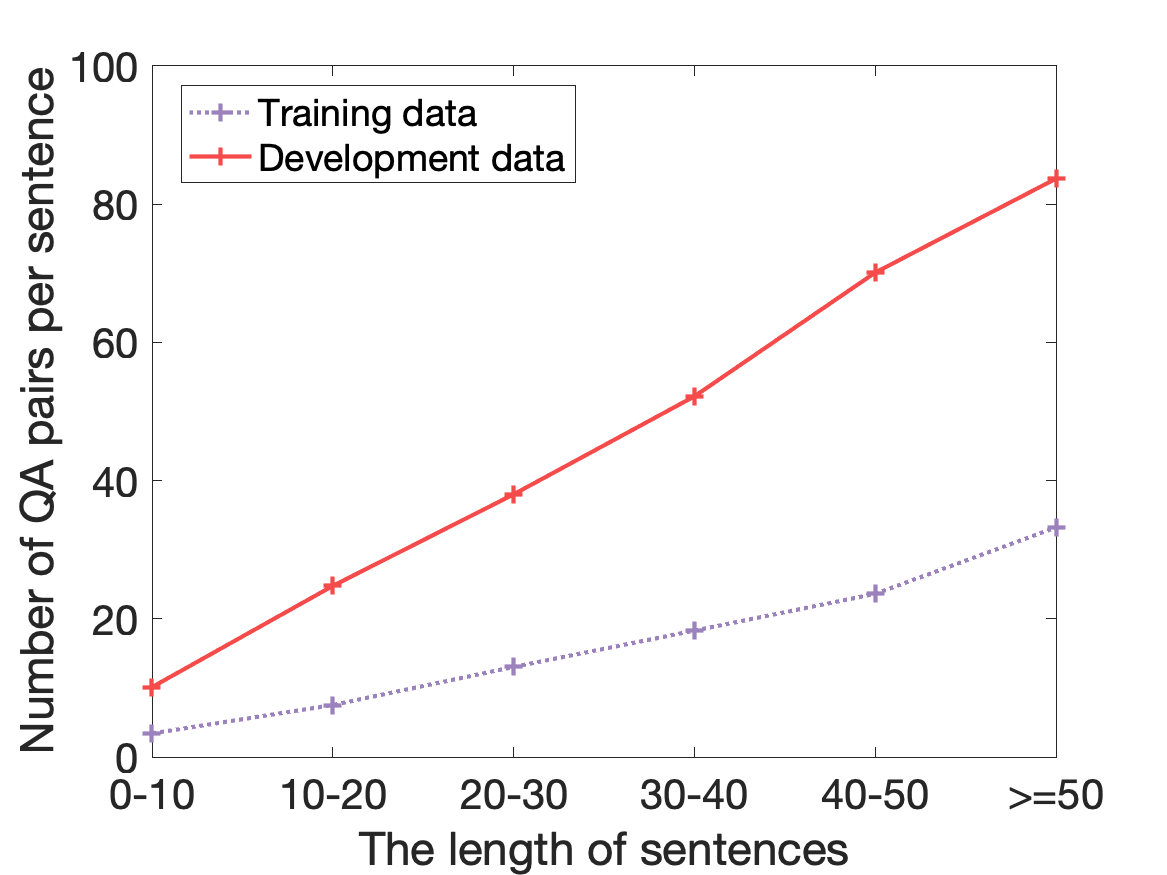}
			\label{fig:error-avg-qa}}
		\caption{
% 		\qn{fontsize too small in figures}
		Error analysis of \name{} on the sentence length. We compare the performance of s-\name{} and p-\name{} on examples with different sentence lengths in the development set. The average number of QA pairs corresponding to the sentence length in the train and development sets is also shown.}
		\label{fig:error-sent}
\end{figure*}
\begin{table*}
\centering
\scalebox{1.0}{
\begin{tabular}{c||c|c|c|c|c}
\Xhline{2\arrayrulewidth}
Models & Model I &  Model II & Model III & Model IV & Model V  \\ \hline
Average EM & 34.97 & 41.64& 55.68 & 64.18 &  {\bf 66.77} \\ \hline
Average F1 & 40.05 & 45.49 & 62.98 & 72.96 & {\bf 76.20} \\ \Xhline{2\arrayrulewidth}
\end{tabular}}
\caption{The results of five variants of s-\quase{QAMR} on the development set of QAMR. We use the average exact match (EM) and average F1 as our evaluation metrics.}
\label{table:modelresults}
\end{table*}
\begin{table}
\centering
\scalebox{0.7}{
\begin{tabular}{c||c|c||c|c||c|c}
\Xhline{2\arrayrulewidth}
Embeddings & \multicolumn{4}{c||}{ELMo} & \multicolumn{2}{c}{Flair} \\ \hline
Tasks & \multicolumn{2}{c||}{SRL} & \multicolumn{2}{c||}{Coref} & \multicolumn{2}{c}{NER} \\ \hline
Split & small & full & small & full & small & full \\ \hline
Baselines & 78.32& 83.87 & 60.72& {\bf 66.89} & 89.86 & 92.37 \\ \hline
s-\quase{QAMR} & {\bf 79.40} & {\bf 84.14} & {\bf 61.54 }& 66.58 & {\bf 90.18} & {\bf 92.54} \\ \Xhline{2\arrayrulewidth}
\end{tabular}}
\caption{Comparison between s-\quase{QAMR} and other STOA embeddings. We use the same experimental settings as Section \ref{subsec:effectiveness-quase} for the three single-sentence tasks, SRL, Coref and NER. We use ELMo embeddings for SRL and Coref, and Flair embeddings for NER as our baselines. }
\label{table:other-embeddings}
\end{table} 
\begin{table*}[t]
\centering
\scalebox{0.86}{
\begin{tabular}{p{102mm}|l|p{36mm}|p{20mm}}
\Xhline{2\arrayrulewidth}
\multicolumn{1}{c|}{Sentence} & \multicolumn{1}{c|}{Ann.} & \multicolumn{1}{c|}{Question} & \multicolumn{1}{c}{Answers} \\ \hline
% Sentence & Ann. & Question & Answers \\ \hline
 (1) {\small Mr. Agnew was vice president of the U.S. from 1969 until he {\bf resigned} in 1973 .} & INF & {\small What did someone resign from?} & {\small vice president of the U.S.}  \\ \hline
 (2) {\small This year , Mr. Wathen says the firm will be able to {\bf service} debt and still turn a modest profit .} & INF & {\small When will something be serviced?} & {\small this year}  \\ \hline
  (3) {\small Mahlunga has said he did nothing wrong and Judge Horn said he "failed to express genuine remorse".} & INF & {\small Who doubted his remorse was genuine?} & {\small Judge Horn}  \\ \hline
  (4) {\small Volunteers are presently renovating the former post office in the town of Edwards, Mississippi, United States for the doctor to have an office.} & IMP & {\small What country are the volunteers renovating in?} & {\small United States}  \\ \Xhline{2\arrayrulewidth}
\end{tabular}}
\caption{Some examples of question-answer pairs in QA-SRL and QAMR datasets. The first two examples are from QA-SRL dataset and predicates are bolded. The last two examples are from QAMR dataset. We show two phenomena that are not modeled by traditional symbolic representations of predicate-argument structure (e.g SRL and AMR), inferred relations (INF) and implicit arguments (IMP).}
\label{table:examples}
\end{table*}
\begin{table*}
\centering
\scalebox{0.76}{
\begin{tabular}{c||c|c|c||c|c|c||c|c|c}
\Xhline{2\arrayrulewidth}
 & \multicolumn{3}{c||}{Span} & \multicolumn{3}{c||}{IOU $\geq 0.5$} & \multicolumn{3}{c}{Token} \\ \hline
 Models & Precision & Recall & F1 & Precision & Recall & F1 & Precision &Recall & F1 \\ \hline 
 Rules + EM &24.31 & 22.78  & 23.52 & 34.34 &32.27 & 33.27 &  50.46 &28.19  & 36.17 \\ \hline 
 PerArgument + CoDL + Multitask & 32.02& 12.30  &17.77  &46.99  &18.06 & 26.09  &70.76 & 17.80 &28.45 \\ \hline 
 Argument Detector + Argument Classifier &  {\bf 49.19}&  {\bf 43.09} & {\bf 45.94} & {\bf 57.84}  & {\bf 50.82} & {\bf 54.10}  & {\bf 69.37} & {\bf 47.60} & {\bf 56.45} \\ \hline  \hline
Mapping: upper-bound &67.82 & 48.58  & 56.61 & 89.09 & 65.82 &75.70  & 91.57 & 70.25 &79.50  \\ \Xhline{2\arrayrulewidth}
\end{tabular}}
\caption{Results of learning an SRL parser from question-answer pairs. }
\label{table:srl-qa-all}
\end{table*}

%\begin{table}[t]
%\centering
%\scalebox{0.9}{
%\begin{tabular}{c||c|c|c|c}
%\Xhline{2\arrayrulewidth}
%Tasks & \multicolumn{4}{|c||}{TE} & \multicolumn{4}{|c|}{MRC} \\ \hline
%Tasks  & \multicolumn{2}{c|}{NER} & \multicolumn{2}{c}{Sentiment} \\ %\hline
%split & 10\% & 100\% & 10\% & 100\% \\ \hline
%BERT  &84.56 &94.35 &91.17 &92.78 \\ \hline
%p-\quase{QAMR} & 86.88 & 94.02 & 91.97 & 92.89 \\ \hline
%Improvement &2.32 &-0.33 & 0.80 &0.11 \\ \Xhline{2\arrayrulewidth}
%\end{tabular}}
%\caption{The results of p-\quase{QAMR} on two single-sentence tasks: NER and sentiment analysis. We use the same experimental settings as Section \ref{subsec:effectiveness-quase}. For simplicity, we use the uncased NER for BERT fine-tuning, because our p-\quase{QAMR} is uncased.}
%\label{table:p-quase-single}
%\end{table}

\section{Additional Details for \name{}}
\label{sec:quase-details}
%\qn{remember to submit the appendix as a separate file; submitting it with the main pdf will lead to rejection.}
In this section, we show the experimental details of \name{}. We first show the experimental settings of training p-\name{} and s-\name{} in Section \ref{subsec:quase-experiments}. After that, we conduct error analysis of \name{} to show the shortcomings of \name{} in Section \ref{subsec:erro-analysis}. Finally, the ablation analysis of s-\name{} is in Section \ref{subsec:quase-ablation}.

\subsection{Experimental Settings}
\label{subsec:quase-experiments}
 Our \name{} is based on the re-implementation of BERT with pytorch \cite{Wolf2019HuggingFacesTS}. Although we might change a bit to fit the memory of GPU sometimes, the common hyper parameters for further pre-training s-\name{} and p-\name{} are as follows:
 
{\bf Further pre-training p-\name{}.} For sentence-level QA datasets (QAMR, Large QA-SRL, and QA-RE), we further pre-train BERT for $4$ epochs with a learning rate of $5e$-$5$, a batch size of $32$, a maximum sequence length of $128$. For paragraph-level QA datasets (SQuAD, TrivaQA, and NewsQA), we further pre-train BERT for $4$ epochs with a learning rate of $5e$-$5$, a batch size of $16$, a maximum sequence length of $384$. 

{\bf Further pre-training s-\name{}. } For sentence-level QA datasets (QAMR, Large QA-SRL, and QA-RE), we further pre-train s-\name{} for $64$ epochs with a learning rate of $1e$-$4$, a batch size of $72$, a maximum sentence length of $128$ and a maximum question length of $24$. For paragraph-level QA datasets (SQuAD, TrivaQA, and NewsQA), we further pre-train s-\name{} for $32$ epochs with a learning rate of $1e$-$4$, a batch size of $8$, a maximum sentence of $384$, and a maximum question length of $64$. We need to note that s-\name{} contains more architectures than BERT, so the hyper parameters for BERT fine-tuning are not good for s-\name{} further pre-training
\subsection{Error Analysis of \name{}}
\label{subsec:erro-analysis}

%\qn{is your purpose here purely to explain why our performance on SRL isn't SOTA? if that's the case, move this section to the appendix and add a footnote somewhere saying that we're aware of the inferior performance, and we know why, and this doesn't change our story, etc.}

The F1 scores of s-\quase{QAMR} and p-\quase{QAMR} on the development set are $76.20$ and $90.35$. In general, the results of s-\name{} are similar to BiDAF \cite{seo2016bidirectional} but are significantly worse than the p-\name{} on QAMR. We conduct thorough error analysis including: sentence length, answer length, question length, question words and the PoS tag of the answer. We find that s-\name{} is not good at dealing with long sentences compared to p-\name{}. The analysis of model performance with regard to sentence length is shown in Figure \ref{fig:error-sent-f1}. The average number of QA pairs is much larger when the sentence is longer as shown in Figure \ref{fig:error-avg-qa}. The distribution of training set and development set is quite different, which makes the situation more complicated. We further compare s-\name$_{Large\;QA-SRL}$ and p-\name{}$_{Large\;QA-SRL}$ on Large QA-SRL whose distribution of training and development sets are the same. From the results, s-\name{} is still not as good as p-\name{} on long sentences. We think that the failure of s-\name{} in long sentences is mainly because there are more relations to encode, while p-\name{} only needs to encode information based on specific questions. We believe that there is a trade-off between the quality and the quantity of sentence information that a model can encode in practice, although $h(S)$ also include the information in $h_A(S)$ in a perfect world.

\subsection{Ablation Analysis for s-\name{}}
\label{subsec:quase-ablation}

s-\name{} consists of three basic components: a sentence encoder for the sentence representation, a question encoder for the question representation, an interaction layer between the sentence component and the question component. We carefully designed five variants of s-\name{} with increasing complexity and performance: (I) Basic model: a fixed BERT and one-layer bidirectional transformer for sentence modeling, a fixed BERT and one-layer bidirectional transformer for question modeling, and a two-layer multi-layer perceptron (MLP) for the interaction layer; (II) a fine-tuned BERT; (III) the same as model II, with a bi-directional attention flow added to the question component; (IV) the same as model III, with the interaction layer changed from a two-layer MLP to a bidirectional transformer; (V) the same as model IV, with the sentence modeling layer and question modeling layer changed from a single-layer bi-directional transformer to a two-layer one, and beam search is used in the inference stage. Table~\ref{table:modelresults} shows the results of our models further pre-trained on the development set of the QAMR dataset. 
%\qn{what's the next sentence?}To some extend, the selection of each component of s-\name{} is meaningful.

\section{Detailed Experimental Setup}
\label{sec:experiment-details}
In this section, we show the details of experimental setup in Section \ref{sec:experiments}. Because the corresponding settings are too many, we show some common settings here and more details are in our code. We first show the details of simple models in Section \ref{subsec:simple-models}, and then show some common  experimental settings of \name{} in Section \ref{subsec:main-experimental-settings}. Finally, we compare s-\name{} with other SOTA embeddings (ELMo and Flair) in Section \ref{subsec:other-embeddings}

\subsection{Simple Models}
\label{subsec:simple-models}
When \name{} is used in the feature-based approach, we need use models for the tasks. For simplicity, we sometimes choose to use some simple models rather than strong baselines in Section \ref{sec:experiments} in our analysis. Following standard practice, we use a simple BiLSTM model with the input of word embeddings and binary features of predicates for SRL, a simple biaffine model based on BiLSTM for SDP, a simple BiLSTM mode for NER, a simple CNN baseline with the input of word embeddings and position features for RE, and a simple BiLSTM model for TE. 

\subsection{Experimental Settings}
\label{subsec:main-experimental-settings}
We use the re-implementation of SRL, NER and Coref from AllenNLP \cite{Gardner2017AllenNLP} for strong baselines, and we implement the strong baselines of SDP and RE ourselves. As for MRC and TE, we use the re-implementation of BERT with pytorch \cite{Wolf2019HuggingFacesTS}. As for simple models, we implement them by ourselves. As for the hyper parameters for strong baselines of single-sentence tasks, we use the same hyper parameters in the related papers (shown in Section \ref{sec:experiments}). As for the hyper parameters for simple models, we tune them ourselves to find some reasonable hyper parameters. The hyper parameters of MRC and TE for p-\name{} are based on \cite{Wolf2019HuggingFacesTS}.

\subsection{Comparison with Other Embeddings}
\label{subsec:other-embeddings}

To show whether s-\name{} can also provide extra features than other STOA embeddings\footnote{The reported STOA models for SRL and Coref is based on ELMo embeddings and the reported STOA model for NER is based on Flair embeddings.}, such as ELMo and Flair, we compare s-\quase{QAMR} embeddings with ELMo embeddings on SRL and Coref, and compare s-\quase{QAMR} embeddings with Flair on NER. The results are shown in Table \ref{table:other-embeddings}. We find that s-\quase{QAMR} has a better performance than ELMo and Flair, especially in the low-resource setting, which indicates that s-\name{} can provide extra features than ELMo and Flair.

\section{On the Strength of Distributed Meaning Representations}
In this section, we first show more details of the comparison between \name{} with symbolic meaning representations in Section \ref{subsec:symbol-representations-details}. After that, we show the details of learning an SRL parser from QA-SRL in Section \ref{subsec:srl-qa-details}. 

\subsection{Comparison with Symbolic Meaning Representations}
\label{subsec:symbol-representations-details}

{\bf Probing Analysis.} We first show the details of our probing analysis on the Xinhua subset\footnote{Only four subsets in AMR dataset contain both training and development sets, but the other three subsets either use informal languages or templatic and report-like structures, which are quite different from the domain of QAMR.} of AMR dataset. Our probing task can be formulated as follows: given two nodes in order, the probing model needs to predict the directed relation from one node to the other. We only consider the cases where there is indeed a relation between them. 

There are $741$ sentences and $9008$ relations in valid alignments with $70$ different types of relations in the training set, and $99$ sentences with $1098$ relations in valid alignments with $43$ different types of relations in the development set. We use the same edge probing model as \cite{tenney2019you}, but we train it by minimizing a softmax loss rather than binary cross-entropy loss. Therefore, our probing results are based on the classification accuracy, not binary F1 score.  

{\bf Systematic Analysis.} We use Large QA-SRL as a testbed to analyze the \textbf{representation ability} of p-\quase{QAMR}. Our p-\quase{QAMR} achieves $85.79$ F1 score on the development set of Large QA-SRL, while BERT further pre-trained on SQuAD with the same number of QA pairs only achieves an F1 score of $64.63$ (it achieves $86.98$ F1 on SQuAD). For reference, BERT further pre-trained on Large QA-SRL can achieve $92.19$ F1 on Large QA-SRL. All these numbers indicate that p-\quase{QAMR} has a strong ability to answer questions related to SRL. 

On the other hand, BERT further pre-trained on Large QA-SRL can only achieve $72.17$ F1 on the development set of QAMR, while p-\quase{QAMR} can achieve $85.79$ F1  on Large QA-SRL (it achieves $90.35$ F1 on QAMR). These results show that QAMR can cover the questions related to SRL, but Large QA-SRL cannot cover many questions related to AMR. Therefore, \textbf{QAMR is a good choice for \name{} to be further pre-trained on}.

{\bf Some Examples.} He et al.~\shortcite{HeLeZe15} show that QA pairs in QA-SRL often contain inferred relations, especially for why, when and where questions. These inferred relations are typically correct, but outside the scope of PropBank annotations \cite{kingsbury2002treebank}. This indicates that QA-SRL contains some extra information about predicates. Some examples are shown in Table \ref{table:examples}.  We further verify that p-\quase{QAMR} can correctly answer questions in the examples, which means that \name{} can encode some extra information that SRL cannot.

Michael et al.~\shortcite{MSHDZ17} show that QAMR can capture a variety of phenomena that are not modeled in traditional representations of predicate-argument structure, including instances of co-reference, implicit and inferred arguments, and implicit relations (for example, between nouns and their modifiers). Some examples of QAMR are shown in Table \ref{table:examples}. Similar to SRL, we find that p-\name{} precedes traditional representations, such as AMR, by correctly answering questions in the examples and hence encoding extra information. 

\subsection{Learning an SRL Parser from QA-SRL}
\label{subsec:srl-qa-details}
\subsubsection{Learng an SRL Parser}
We consider learning a SRL parser from QA-SRL. It reduces the problem of learning AMR from QAMR to a simplified case.

{\bf Challenges.}
There are three main challenges to learn an SRL parser from Large QA-SRL.
%\begin{itemize}
 %\item 
 
 \textit{Partial issues.} Only $78\%$ of the arguments have overlapped with answers; $47\%$ of the arguments are exact match; $65\%$ of the arguments have Intersection/Union $\geq$ $0.5$\footnote{These statistics of partial issues and irrelevant question-answer pairs are based on the PTB set of QA-SRL.}.
 
 %\item 
 \textit{Irrelevant question-answer pairs.} $89\%$ of the answers are ``covered" by SRL arguments; $54\%$ of the answers are exact match with arguments; $73\%$ of the answers have Intersection/Union $\geq$ $0.5$. These statistics show that we also get some irrelevant signals: some of the answers are not really arguments (for the corresponding predicate).
 
 %\item 
 \textit{Different guidelines.} Even if the arguments and the answer overlap, the overlap is only partial. 
 
 %\item 
 %\textit{Cross domain issues.} We need to evaluate the trained SRL model on Propbank, but corresponding QA pairs are not annotated in Propbank dataset. For example, Large QA-SRL annotates sentences in three domains: Wikipedia, Wikinews and Science.\\
%\end{itemize}
{\bf A reasonable upperbound.} We treat the answers that have overlapped with some arguments as our predicted arguments. If two predicted arguments intersect each other, we will use the union of them as new predicted arguments. The results are shown in Table \ref{table:srl-qa-all}. We know from the table that this mapping algorithm achieves a span F1 of $56.61$, which is a reasonable upper bound of our SRL system.

{\bf Baselines.} We consider three models to learn an SRL parser from Large QA-SRL dataset.

%\begin{itemize}
    %\item 
    \textit{Rules + EM.} We first use rules to change QA pairs to labels of SRL. We keep the labels with high precision and then use an EM algorithm to do bootstrapping. A simple BiLSTM is used as our model for SRL. The results are shown in Table \ref{table:srl-qa-all}. We think that low token F1 is due to the low partial rate of tokens ($37.97\%$) after initialization.
    
    %\item 
    \textit{PerArgument + CoDL + Multitask.} We consider a simpler setting here. A small number of gold SRL annotations are provided as seeds. To alleviate the negative impact of low partial rate, we propose to train different BiLSTM models for different arguments (PerArgument) and do global inference to get structured predictions\footnote{Given a predicate in the sentence with three arguments and one of them is annotated, the sentence is partial for a traditional SRL model but not partial for a PerArgument model.}. We first use seeds to train the PerArgument model and then use CoDL \cite{ChangRaRo07} to introduce constraints, such as SRL constraints, into bootstrapping. At the same time, we train a model to predict the argument type from question-answer pairs. These two tasks (argument type prediction and SRL) are learned together through soft parameter sharing. In this way, we make use of the information from QA pairs for SRL. We use $500$ seeds to bootstrap. The span F1 of our method is $17.77$ and the span F1 with only seeds is $13.65$. More details are in Table \ref{table:srl-qa-all}. The performance of this model has only improved several percents compared to the model trained only on seeds.
    
    %\item 
    \textit{Argument Detector + Argument Classifier.} Given a small number of gold SRL annotations and a large number of QA pairs, there are two methods to learn an end-to-end SRL\footnote{Note that an end-to-end SRL system is with gold predicates. This is different from the generic definition.} system. One is to assign argument types to answers in the context of corresponding questions using rules, and learn an end-to-end SRL model based on the predicted SRL data. This is exactly our first model, Rules + EM. However, the poor precision of argument classification leads to unsatisfactory results. Another method is to learn from small seeds and bootstrap from large number of QA pairs. Thich is our second model, PerArgument + CoDL + Multitask. However, bootstrapping can not improve argument detection much, leading to mediocre results. We also notice that argument detection is hard with a small number of annotated data, but argument classification is easy with little high-quality annotated data. Fortunately, most answers in Large QA-SRL overlap with arguments. Furthermore, the mapping results of argument detection is about $56.61$, good enough compared to two baselines. We propose to learn two components for SRL, one is for argument detection and the other is for argument classifier. We use the span-based model in \cite{fitzgerald-etal-2018-large} for argument detection. The argument classifier is trained on predicates in the PTB set of QA-SRL. The results are shown in Table \ref{table:srl-qa-all}.
%\end{itemize}
\subsubsection{Using SRL/AMR Parsers in Downstream Tasks}
There have already been some attempts to use semantics in downstream tasks. We discuss three types of application here. Traditionally, semantic parsers can be used to extract semantic abstractions, and can be applied to question answering \cite{KKSR18}. Second, dependency graphs, such as SDP, can be incorporated into neural networks. For example, \cite{marcheggiani-titov-2017-encoding} encodes semantic information in Graph Convolution Networks (GCN). In order to use constituent based traditional symbolic meaning representations, one can encode related semantic information by multi-task learning (MTL). \cite{strubell-etal-2018-linguistically} mentioned such an example of application.

The main difficulty of retrieving SRL/AMR from QA signals for downstream tasks is to learn a good parser for SRL/AMR from question-answer pairs.

\end{document}